\documentclass[10pt,journal,compsoc]{IEEEtran}
\ifCLASSOPTIONcompsoc
  \usepackage[nocompress]{cite}
\else
  \usepackage{cite}
\fi
\ifCLASSINFOpdf
  \usepackage[pdftex]{graphicx}
  \DeclareGraphicsExtensions{.pdf,.jpeg,.png}
\else
\fi
\usepackage{float}
\usepackage{subfig}
\usepackage[flushleft]{threeparttable}
\usepackage{adjustbox}
\usepackage{tablefootnote}
\usepackage[font={footnotesize}]{caption}
\usepackage{hyperref}
\usepackage{tablefootnote}
\usepackage{amsmath}
\usepackage{xcolor} 
\usepackage{dblfloatfix} 
\usepackage{caption}
\usepackage{booktabs,array}

\usepackage{algorithmic}
\begin{document}
%
\title{Universal 3D Wearable Fingerprint Targets: Advancing Fingerprint Reader Evaluations}


%
%
%
%

\author{Joshua J.~Engelsma,
        Sunpreet S.~Arora,~\IEEEmembership{Member,~IEEE,} \\Anil K.~Jain,~\IEEEmembership{Life~Fellow,~IEEE,}
        and~Nicholas G.~Paulter Jr.,~\IEEEmembership{Fellow,~IEEE}
\IEEEcompsocitemizethanks{\IEEEcompsocthanksitem J. J. Engelsma and A. K. Jain are with the Department of Computer Science and Engineering, Michigan State University, East Lansing, MI, 48824\protect\\
E-mail: \{engelsm7, jain\}@cse.msu.edu

\IEEEcompsocthanksitem S. S. Arora was with the Department of Computer Science and Engineering, Michigan State University. He is now with the Risk and Authentication Products organization at Visa Inc., Foster City, CA 94404\protect\\
Email: sunarora@visa.com

\IEEEcompsocthanksitem N. G. Paulter Jr. is with the National Institute of Standards and Technology (NIST), Gaithersburg, Maryland 20899\protect\\
Email: paulter@nist.gov

}

}

\IEEEtitleabstractindextext{%
\begin{abstract}
We present the design and manufacturing of high fidelity universal 3D fingerprint targets, which can be imaged on a variety of fingerprint sensing technologies, namely capacitive, contact-optical, and contactless-optical. Universal 3D fingerprint targets enable, for the first time, not only a repeatable and controlled evaluation of fingerprint readers, but also the ability to conduct fingerprint reader interoperability studies. Fingerprint reader interoperability refers to how robust fingerprint recognition systems are to variations in the images acquired by different types of fingerprint readers. To build universal 3D fingerprint targets, we adopt a molding and casting framework consisting of (i) digital mapping of fingerprint images to a negative mold, (ii) CAD modeling a scaffolding system to hold the negative mold, (iii) fabricating the mold and scaffolding system with a high resolution 3D printer, (iv) producing or mixing a material with similar electrical, optical, and mechanical properties to that of the human finger, and (v) fabricating a 3D fingerprint target using controlled casting. Our experiments conducted with PIV and Appendix F certified optical (contact and contactless) and capacitive fingerprint readers demonstrate the usefulness of universal 3D fingerprint targets for controlled and repeatable fingerprint reader evaluations and also fingerprint reader interoperability studies.
\end{abstract}

}

\maketitle

\IEEEdisplaynontitleabstractindextext

%
\IEEEpeerreviewmaketitle

\IEEEraisesectionheading{\section{Introduction}\label{sec:introduction}}

%
%
%
%

\IEEEPARstart{A}{utomated} fingerprint identification systems (AFIS) have become increasingly ubiquitous over the last fifty years. With origins in the forensics community in the early 1900s, fingerprints have continued to serve as valuable links to individuals due to their proven uniqueness, permanence, universality, and collectability \cite{handbook}. More recently, fingerprint recognition systems have exploded into a plethora of niche areas such as mobile device security, healthcare access, financial systems, and government institutions \cite{handbook}. As fingerprints continue to become a key to access society's confidential data, social benefits, networks, and buildings, the need to know and quantify fingerprint recognition accuracy is paramount. As such, controlled, repeatable evaluations of the various components of fingerprint recognition systems must be performed. While past end-to-end evaluations such as FpVTE 2012 \cite{nist} have provided us with baseline statistics on the performance of state-of-the-art fingerprint recognition systems, much work remains to be done in developing rigorous evaluations of the reader\footnote{A distinction is made between fingerprint reader and fingerprint sensor. Fingerprint reader refers to the entire device and process, which captures your physical fingerprint and converts it into a digital image. The sensor is a subcomponent of the reader which converts, through a variety of means (capacitive, frustrated total internal reflection), the physical fingerprint to an electrical signal.} subcomponent of fingerprint recognition systems.

\begin{figure}[!t]
\begin{center}
\includegraphics[scale=0.15]{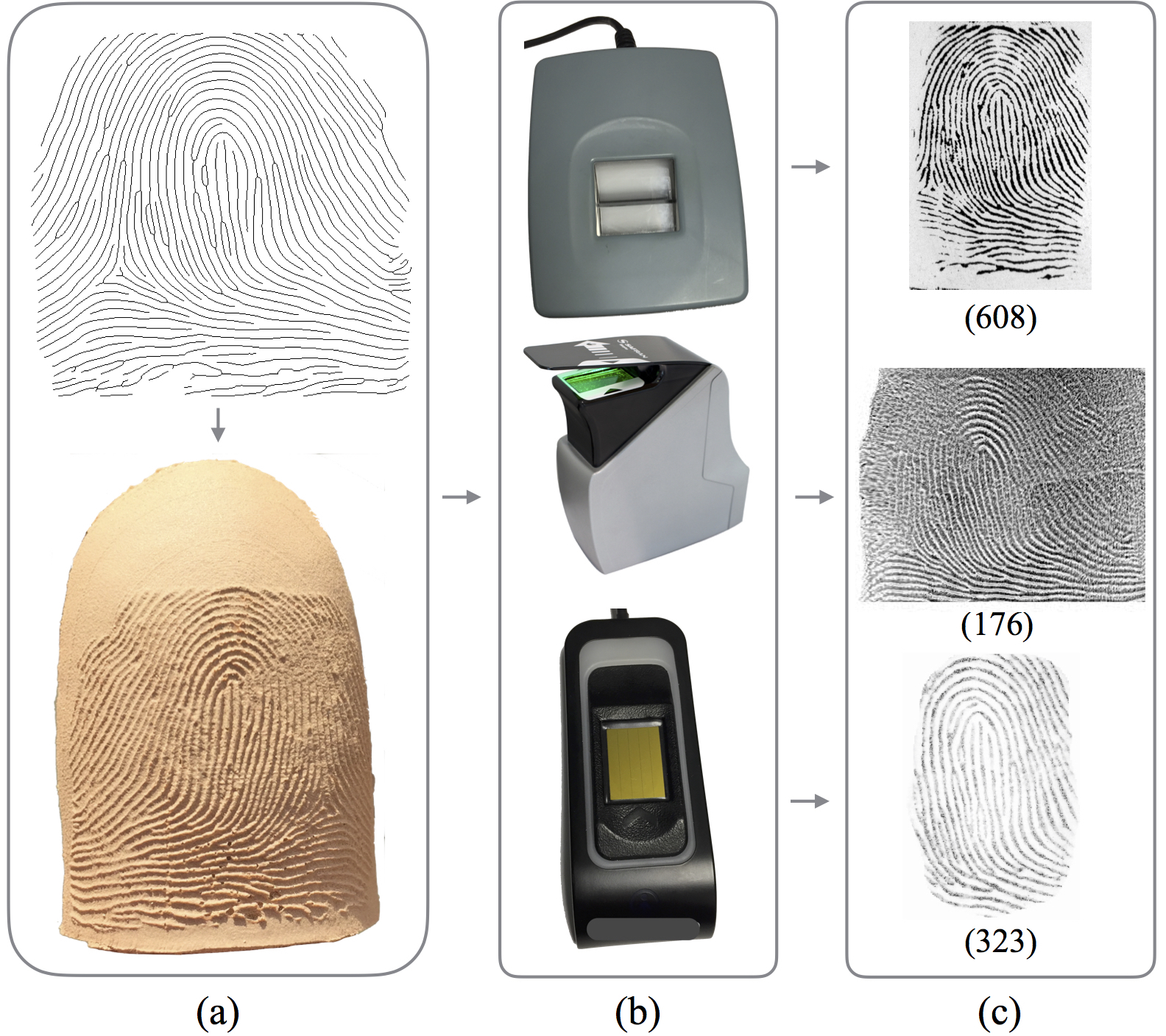}
\caption{A Universal 3D Fingerprint Target fabricated in (a) can be imaged by a variety of popular fingerprint readers (contact-optical, contactless-optical, and capacitive) shown in (b). The sensed images of the 3D fingerprint target in (a) are shown in (c). This demonstrates that our targets are appropriate for fingerprint reader interoperability evaluation studies. Similarity scores for each sensed fingerprint image (with the 2D mapped target image) are displayed below each fingerprint image in (c). Verifinger 6.3 SDK was used for generating similarity scores. The score threshold at 0.01 \% FAR is 33.}
\end{center}
\vspace{-1.0em}
\end{figure} 

\begin{figure*}[t]
\begin{center}
\includegraphics[scale=0.25]{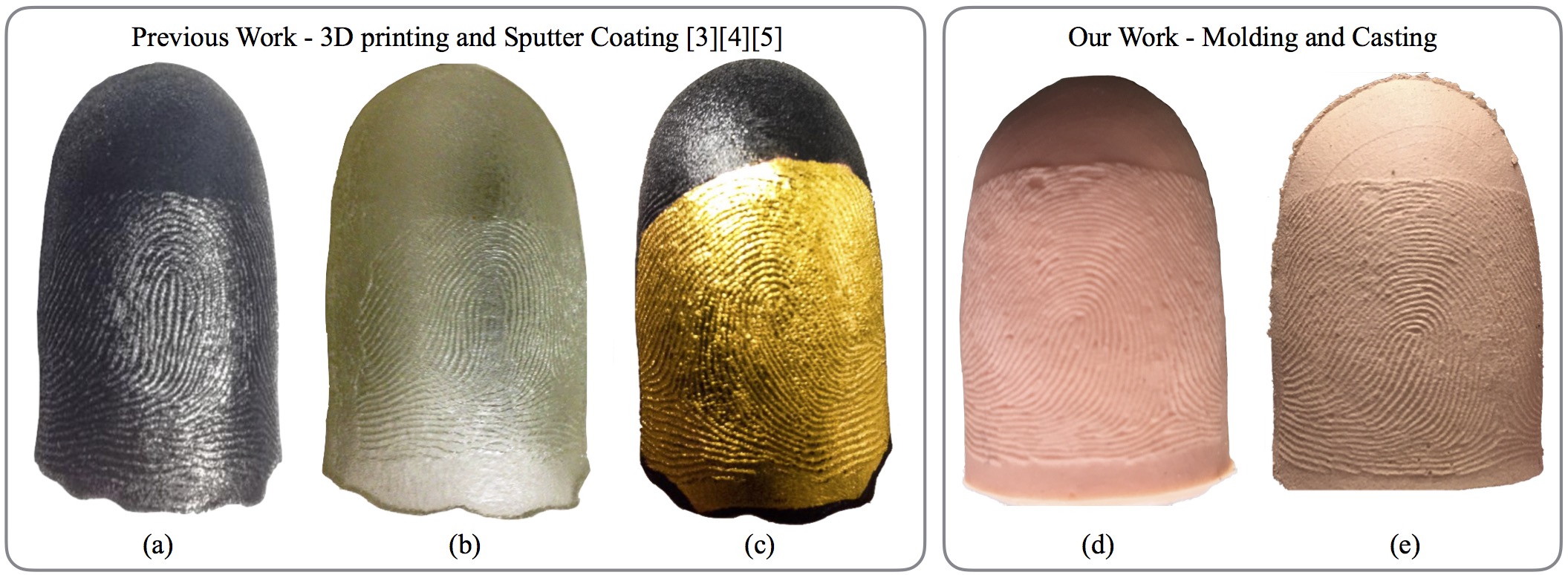}
\caption{High fidelity, wearable, 3D fingerprint targets. (a) 3D fingerprint target printed using TangoBlackPlus FLX980 \cite{3Dfingers}, (b) 3D fingerprint target printed using TangoPlus FLX 930 \cite{whole_hand}, (c) 3D fingerprint target printed using TangoBlackPlus FLX980 and then sputter coated with 30 nm titanium + 300 nm of gold \cite{goldfinger}, (d) our casted 3D fingerprint target using mixture of PDMS (Polydimethylsiloxane) and Pantone 488C color pigment \cite{pigments}\cite{pdms}, and (e) our casted universal 3D fingerprint target using mixture of conductive PDMS, silicone thinner, and Pantone 488C color pigment \cite{pigments}\cite{ss-27s}\cite{thinner}. 3D targets in (a), (b), and (c) were printed on a high resolution 3D printer (Stratasys Objet350 Connex).}
\end{center}
\end{figure*}

\newcommand{\specialcell}[2][c]{%
  \begin{tabular}[#1]{@{}c@{}}#2\end{tabular}}

 \begin{table*}[t]
 \centering
\caption{A Comparison of the Properties of the Human Finger, 3D Printed Targets, and our 3D Casted Targets}
\begin{threeparttable}
\resizebox{\textwidth}{!}{%
\begin{tabular}{ |c||c|c|c|c|c|c|}
 \hline
Specimen Material &Shore A Hardness&\specialcell{Tensile Strength \\(MPa)}&\specialcell{Elongation at Break \\(\%)}&Color&\specialcell{Electrical Resistance \\ (M$\Omega$)} &\specialcell{Cost/Target \\ (USD)}\\
\hline
\hline
 Human Skin \cite{bio1}\cite{bio2} & 20-41 & 5-30 & 35-115 & Varies & 1-2 & N.A.\\
  \hline
 \hline
 \specialcell{TangoBlackPlus FLX980 \\(Fig. 2 (a)) \cite{3Dfingers}\cite{tangoblack}} & 26-28  & 0.8-1.5 & 170-220 & Black & Insulator & \$10.00\\
  \hline
 \specialcell{TangoPlus FLX930 \\(Fig. 2 (b)) \cite{whole_hand}\cite{tangoblack}} & 26-28 & 0.8-1.5 & 170-220 & Translucent & Insulator & \$10.00 \\
  \hline
  \specialcell{TangoBlackPlus FLX980, \\ Ti-Au surface coating (Fig. 2 (c))\\ \cite{goldfinger}\cite{tangoblack}} & 26-28 & 0.8-1.5 &170-220 & Gold & $5*10^{-4}$ & \$12.00 \\
  \hline
  \hline
 \specialcell{PDMS \& Pantone 488C Pigment \\(Fig. 2 (d)) \cite{pdms}} & 43 & 6.7 & 120 & PMS 488C & Insulator & \$0.86\\
  \hline
 \specialcell{Conductive PDMS, Silicone Thinner, \& \\Pantone 488C Pigment \\(Fig. 2 (e)) \cite{pigments}\cite{ss-27s}\cite{thinner}} & 38.5 & \specialcell{2.0} &  \specialcell{80} & Tan / PMS 488C &  $7*10^{-3}$ \tnote{\textdagger} & \$10.00\\
\hline
 
\end{tabular}}
\begin{tablenotes}
\item[\textdagger] Although the resistance of the target differs from human skin, the resistance value is sufficient for image capture by capacitive readers.
\end{tablenotes}
\end{threeparttable}
\end{table*}

\begin{figure}[h]
\begin{center}
\includegraphics[scale=0.175]{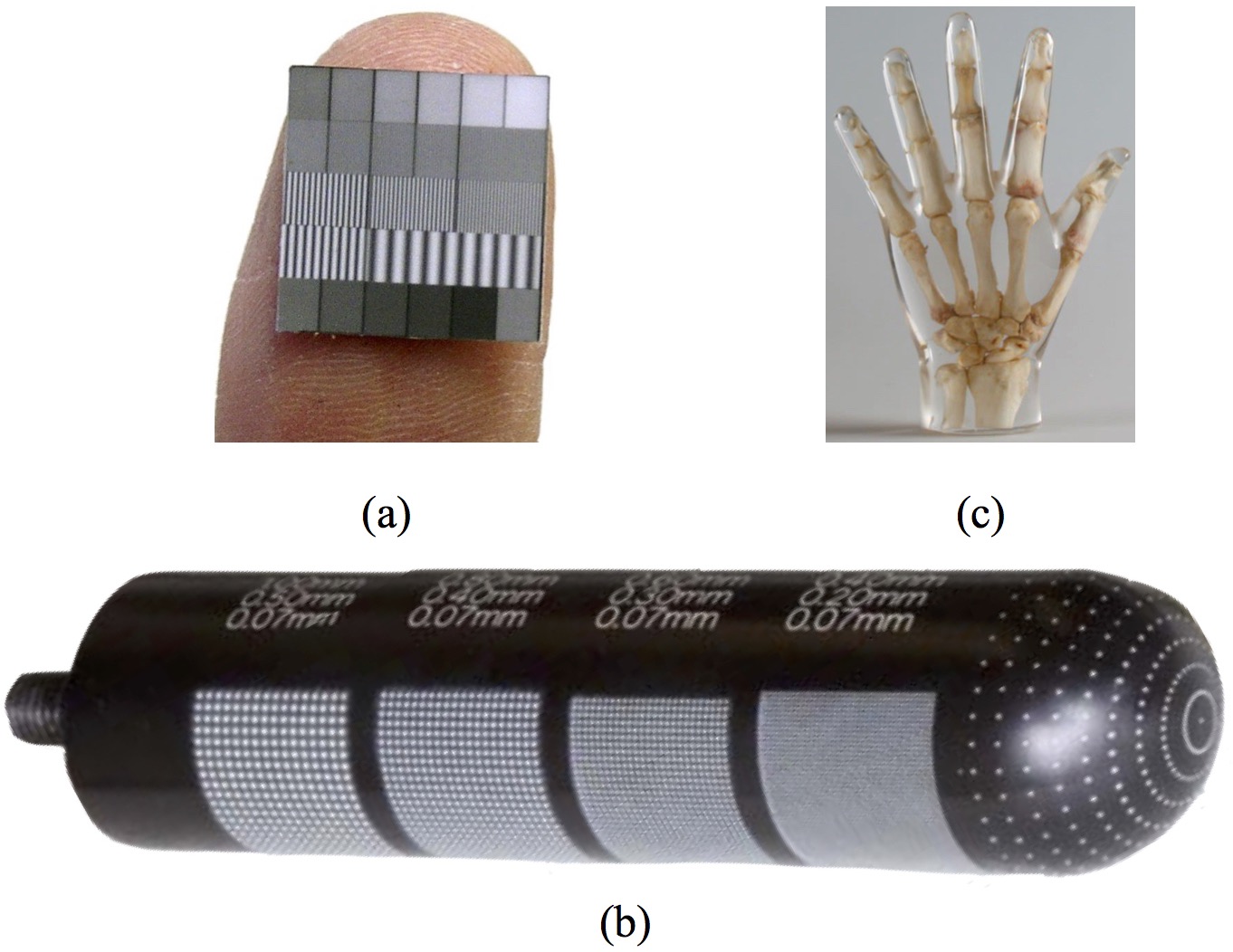}
\caption{Examples of Evaluation Targets. (a) standard 2D fingerprint reader calibration target \cite{2d}; (b) 3D metal cylinder for contactless fingerprint reader calibration \cite{cylinder}; (c) medical phantom of a human hand \cite{phantom1}. Images taken from \cite{2d}\cite{cylinder}\cite{phantom1}}
\vspace{-1.5em}
\end{center}
\end{figure} 

\begin{figure*}[t]
\begin{center}
\includegraphics[scale=0.25]{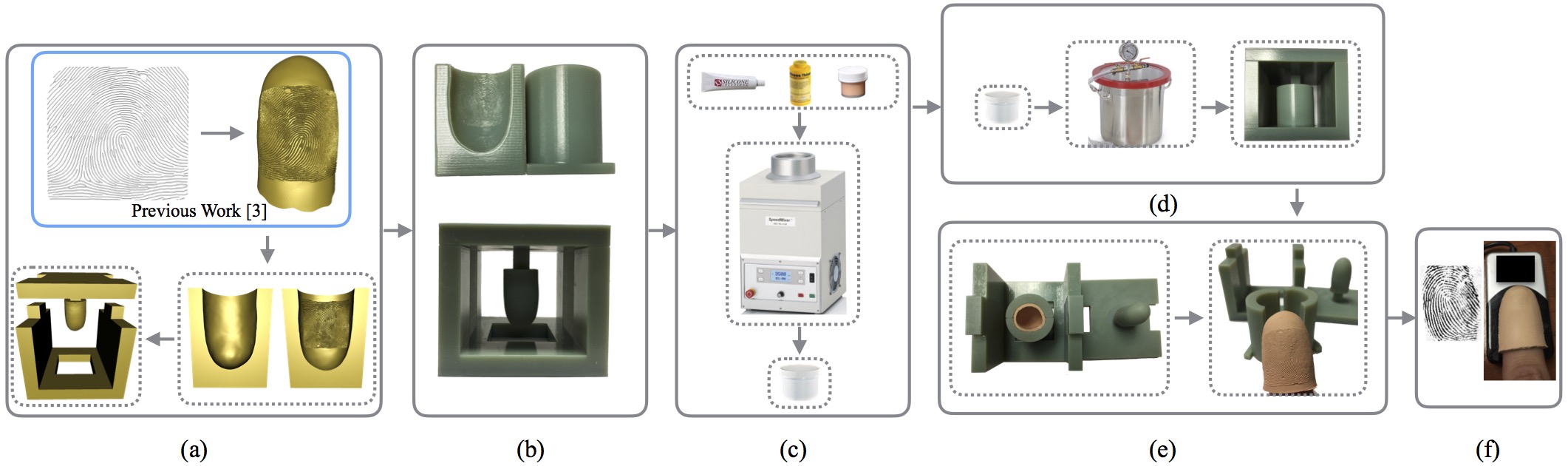}
\caption{System block diagram of the proposed molding and casting process for making 3D targets. (a) A 3D negative mold (of a 2D fingerprint image) and a supporting scaffolding system (necessary for making the fingerprint target wearable) are electronically fabricated; (b) 3D electronic models are manufactured by 3D printing and chemical cleaning; (c) conductive silicone, silicone thinner, and human colored dye are mechanically mixed to produce a casting material with similar conductive, mechanical, and optical properties to the human skin; (d) the material fabricated in (c) is cast into the mold and scaffolding system; (d) vacuum degassing \cite{vacuum} ensures that air bubbles are removed from the casted material; (e) wearable fingerprint targets are extracted 72 hours after pouring the casting material; (f) the wearable, 3D fingerprint target is used for fingerprint reader evaluations.}
\vspace{-1.5em}
\end{center}
\end{figure*}

Previous attempts to evaluate the fingerprint reader component have been predominantly undertaken by the FBI and constitute the {\it Appendix F} and {\it PIV} standards \cite{standards}. The {\it Appendix F} standard is comparatively stringent, requires pristine image capture, and is designed to facilitate evaluation of fingerprint readers used in person identification scenarios (one to many comparisons). The {\it PIV} standard is a softer standard than {\it Appendix F} and is designed to evaluate fingerprint readers used in person verification scenarios (one to one comparison). Both of these standards use imaging targets that are fabricated by projecting a calibration pattern (e.g. sine gratings) onto a flat surface (Fig. 3 (a)) . These targets are useful for structural (white-box)\footnote{White-box testing focuses on testing the internal sub-components of a system, whereas black-box testing focuses on testing the end-to-end system using system inputs and outputs \cite{testing}.} testing of fingerprint readers since they ensure that certain quantitative imaging thresholds are met by the fingerprint reader's sensing component, however, these targets have little resemblance to the human fingers that the readers will be exposed to in an operational setting. As such, controlled operational (black-box) evaluations of fingerprint readers using the existing standards and targets are limited at best. 

To address the challenges of robust operational evaluation inherent to imaging devices, groups from other domains have developed 3D targets (similar to the items which will eventually be imaged) as evaluation specimens. In the medical imaging community, these targets are referred to as phantoms. Phantoms are useful for evaluating a variety of medical imaging devices in areas such as radiography, tomography, and ultrasonic imaging \cite{phantom1}\cite{phantom2}. Use of live subjects for repeated evaluation of radiographic medical devices is impractical because of the health hazards and monetary costs involved.  However, realistic 3D phantoms (Fig 3. (c)) make accurate operational evaluation of these devices possible. We posit that proper operational evaluation of fingerprint readers can only be accomplished, in a similar manner, by using 3D fingerprint targets (phantoms) with similar characteristics to the human finger.

\subsection{3D Fingerprint Targets}

Some research has been conducted developing 3D targets towards achieving the aforementioned goal. In 2011, Orandi et. al developed 3D cylindrical metal targets mapped with 2D calibration patterns for contactless fingerprint readers (Fig. 3 (b)) \cite{nist_patent}. However, because these targets are rigid and completely dissimilar in mechanical, optical, and capacitive properties to the human finger, they can not be used by contact-based fingerprint readers. More recently, in 2016, Arora et. al produced high fidelity 3D fingerprint targets using a high resolution, state-of-the-art 3D printer \cite{3Dfingers}\cite{whole_hand}\cite{goldfinger}. These targets were a big step forward in the direction of realistic operational fingerprint reader evaluation because the targets employed a 3D geometry similar to the human finger, they were fabricated using materials with similar mechanical properties as human skin, they were mapped with real fingerprint images, and they could be worn on a human finger. However, due to the limited number of materials that can be used in 3D printers, the polymers used for printing (i) did not have the same nominal electrical conductivity of human skin and (ii) did not have the spectral reflectance of human skin. As a result, multiple types of targets (Figs. 2 (a), (b), (c)) were fabricated for different types of fingerprint readers (capacitive, contact-optical, and contactless-optical) \cite{3Dfingers}\cite{whole_hand}\cite{goldfinger}. These individual targets worked for evaluating the type of reader for which they were designed, however, they were not interoperable. That is, a target fabricated for one type of fingerprint reader (e.g. capacitive) would not work on a different type of fingerprint reader (e.g. optical). Because multiple types of targets were needed for evaluating different types of readers, performing a standardized interoperability evaluation of fingerprint reader technologies was not possible with these 3D printed targets. 

\subsection{Fingerprint Reader Interoperability}

Past studies on fingerprint reader interoperability have shown that when different fingerprint readers were used for enrollment and identification (or verification), significant losses in recognition accuracy ensued \cite{Ross2004}\cite{ross_inter}\cite{cost_inter}. However, all of these studies were performed on data acquired from live human subjects \cite{msutech1}. As such, variations (finger pressure and orientation; conditions of the finger, e.g. wet or dry) between impressions on the different readers could account for some of the error observed. We posit that in order to truly quantify the effects of interoperability, an interoperable fingerprint target would need to be mounted to a robot gripper and imaged on different readers at the same pressure and orientation.

\begin{figure*}[t]
\begin{center}
\includegraphics[scale=0.25]{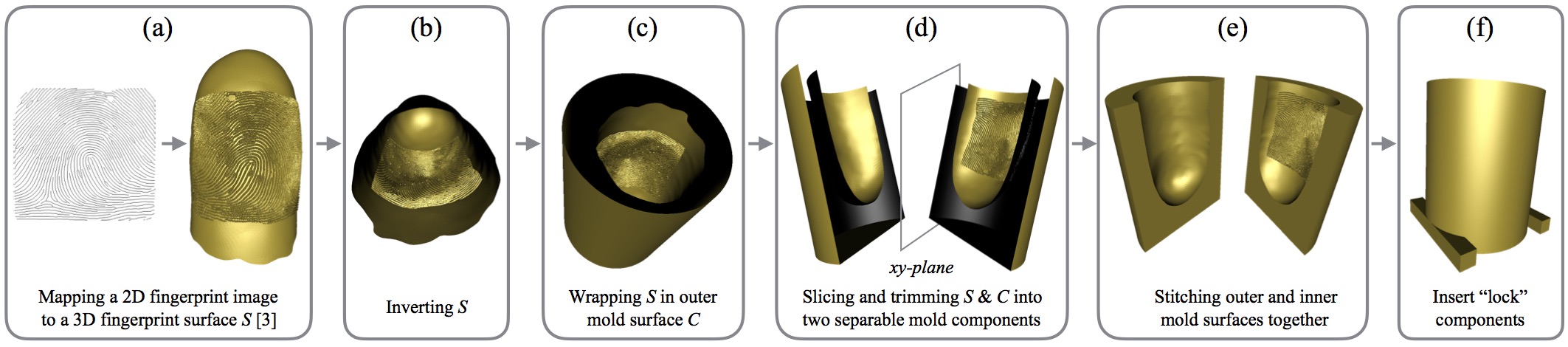}
\caption{Process flow for fabricating electronic 3D fingerprint mold, $M$}
\vspace{-1.0em}
\end{center}
\end{figure*}

As noted in \cite{stats_inter}, continued advances in distributed computing have enabled less monolithic fingerprint recognition systems. This advent of larger, more distributed systems drastically increases the likelihood that the fingerprint reader used to enroll a user's fingerprint image at one location will not be the same reader (or model of reader) used later to identify or verify the same individual at another location. Consider, for instance, India's Aadhaar program, which has already enrolled over 1.14 billion residents (as of May 2017) on a variety of readers, many of whom are receiving services and benefits based on fingerprint and/or iris recognition \cite{india1}\cite{india2}. Furthermore, even if the same reader is used for both enrollment and identification, advances in sensing technology could eventually require replacement of the reader being used. As mentioned in \cite{cost_inter}, the cost to an institution needing to re-enroll its entire database of users on a new reader could be monumental. Both of these situations underscore the need to know and quantify fingerprint reader interoperability. If fingerprint recognition systems are to continue to become more interoperable, then the performance change associated with interoperability must be objectively known and quantified. Doing so will benefit system users, reader manufacturers, system developers, and the institutions deploying the system. 

\subsection{Universal 3D Fingerprint Targets}
To enable robust, standardized fingerprint reader interoperability evaluations, we present the fabrication of an interoperable 3D fingerprint target through a molding and casting process (Fig. 4). We call our target the {\it universal fingerprint target} (Fig. 2 (e)). Like previous fingerprint targets in \cite{3Dfingers}, the universal fingerprint targets share a 3D geometry similar to a fingerprint surface, have mechanical properties similar to human skin, and are mapped with a fingerprint image, either real or synthetic. However, unlike previous fingerprint targets, the universal fingerprint targets are unique in that they incorporate the technically pertinent mechanical, optical, and electrical properties of the human skin within a single target (Table 1), making it possible for the universal fingerprint targets to be imaged by all major fingerprint sensing technologies in use (capacitive, contact-optical, contactless-optical). The universal fingerprint targets enable and facilitate, for the first time, a standardized assessment of fingerprint reader interoperability. The universal fingerprint targets will also enable controlled data collection useful for fingerprint distortion modeling.

More concisely, the contributions of this research are:

\begin{itemize}
 \item A controlled, repeatable process for creating fingerprint target molds, and fabricating high quality finger castings. Unlike previous works \cite{3Dfingers}, this casting fabrication process is not restricted to a small number of materials. Additionally, it is not cost prohibitive as it is based on a potentially high-throughout casting process.
  \item Fabricating high fidelity universal 3D fingerprint targets with similar mechanical, optical, and electrical properties to the human skin. Previous targets did not simultaneously possess both the optical and electrical properties of human skin within a single target.
  \item Fingerprint image capture, using the same 3D target, from optical readers (contact and contactless) and capacitive readers. Our universal fingerprint targets will enable standardized interoperability data collection for the first time ever. 
 \item Experimental evaluations, using the universal 3D fingerprint targets and three different types of commercial off-the-shelf (COTS) fingerprint readers\footnote{Because of our Non-Disclosure Agreement with the vendors, we cannot provide the make and model of the readers used in our experiments.}  (contact-optical, contactless-optical, and capacitive). Our results quantify the loss in fingerprint recognition accuracy when different readers are used for enrollment and identification (or verification). These findings validate the use of our universal 3D fingerprint target for further fingerprint reader interoperability studies.
\end{itemize}

\section{Mold \& Scaffold Fabrication}

To fabricate a fingerprint target $T$, we begin by electronically modeling (and subsequently manufacturing) a fingerprint mold $M$ and scaffolding framework $F$.

\subsection{Mold Fabrication}

First, a negative\footnote{In molding and casting, positive sculptures are produced from their negative mold.} fingerprint mold is electronically designed (Fig. 5), 3D printed, and chemically cleaned. This process is further broken down and expounded upon in the following steps.

{\it i) Inner Mold Surface -} Using techniques similar to \cite{3Dfingers}, a 2D fingerprint image is mapped onto a smooth 3D finger surface mesh $S$ in a manner that retains the topology inherent to the 2D image (Fig. 5 (a)). More formally, let $S$ be a mesh of triangular faces $F = \begin{bmatrix} f_1, & f_2, & f_3, & ..., & f_n \end{bmatrix}$, and 3-dimensional vertices $V = \begin{bmatrix} v_1, & v_2, & v_3, & ..., & v_c \end{bmatrix}$. Each face in $F$ is explicitly defined as an ordered list of 3 vertices from $V,$ e.g. $f_1 = \begin{bmatrix} v_i, & v_j, & v_k \end{bmatrix}$. Additionally, every face in $F$ contains a normal vector which is implicitly encoded by the order of the 3 vertices used to define the face. In particular, the direction of the normal vector is determined by taking the cross product of the vectors formed with respect to the order of the face's three vertices. For example, the normal vector for face $f_1$ is $f_{1,normal} = a \times b$, where $a$ is a vector having tail at $v_i$ and head at $v_j$, while $b$ is a vector having tail at $v_j$ and head at $v_k$.

Because the end goal of the electronic modeling of $M$ is to produce a negative mold, the mapped surface $S$ must be inverted by flipping all the faces of $S$ (Fig. 5 (b)). For every face, this flipping is attained by reversing the order of its three vertices - and consequently the implicitly encoded direction of its normal vector. For example, by changing $f_1 = \begin{bmatrix} v_i, & v_j, & v_k \end{bmatrix}$ to $\hat{f_1} = \begin{bmatrix} v_k, & v_j, & v_i \end{bmatrix}$, the normal vector $\hat{f}_{1,normal}$ computed by  $a \times b$ is reversed in direction, since $a$ is now a vector having tail at $v_k$ and head at $v_j$, while $b$ is a vector having tail at $v_j$ and head at $v_i$. 

{\it ii) Outer Mold Surface -}  After iteratively inverting all $n$ $\begin{bmatrix} f_1, & f_2, & f_3, & ..., & f_ n\end{bmatrix}$ faces, the next step in generating mold $M$ is to imprint the fingerprint surface $S$ inside of an open ended cylindrical surface $C$ (Fig. 5 (c)). Surface $C$ acts as the exterior of the final mold $M$. As such, dimensions for $C$ are determined empirically so as to provide strength and durability to the mold and to prevent usage of excess material. Our experiments show that setting the height of $C$ to $C_{height} = 1.25 * S_{height}$ balances the need for structural support and minimizes material cost for casted targets (here $S_{height}$ is the height of the fingerprint surface $S$). The diameter of the mold ($C_{dia}$) is fixed at 34 mm. While $C_{dia}$ could have been dynamically chosen based upon the diameter of $S$ ($S_{dia}$), we chose a fixed value so that all the molds we print could fit within a single scaffolding framework $F$. We chose 34 mm as a static diameter value, since the 95\textsuperscript{th} percentile of the widest adult finger (the thumb) is 26 mm to 27 mm \cite{hand_anthro}. As such, the minimum thickness ($t_{min}$) of our mold is computed as $t_{min} = 1/2 * (34 - 27)mm = 3.5$ mm. We empirically validated that a mold thickness of $t_{min} \geq 3.5$ mm provides the durability needed for our casting process.

{\it iii) Split Mold -} With the inner and outer surface of the mold in place, we continue the fabrication process by simultaneously splitting $C$ and $S$ along the {\it xy-plane} into $C_{above}$, $S_{above}$, $C_{below}$, and $S_{below}$. Splitting the mold into two semi-cylindrical components will facilitate the extraction of the final fingerprint castings $T$ (from the mold). $C_{above}$, $S_{above}$, $C_{below}$, and $S_{below}$ are further post processed by adding new faces and vertices such that all four surfaces lie flat on the xy-plane.  Figure 5 (d) illustrates the sliced, trimmed, and post processed components $C_{below}$, $C_{above}$, $S_{below}$, and $S_{above}$. 

{\it iv) Stitching and Printing -} Finally, the individual surfaces $C_{below}$ and $S_{below}$ and $C_{above}$ and $S_{above}$ are stitched together into two three-dimensional, semi-cylindrical mold halves by adding triangular faces around the periphery of the respective surfaces. Upon completion of this stitching, a high fidelity fingerprint mold $M$ has been electronically fabricated (Fig. 5 (e)). 

To minimize the variability of fingerprint targets during consecutive castings, two {\it ``lock"} components are attached to the bottom of $C$ (Fig. 5 (f)). These lock pieces, having length equal to 34 mm ($C_{dia}$) will prevent $C$ from rotating inside of the scaffolding framework $F$.

At this point, $M$ is physically realized by using a high resolution, state-of-the-art 3D printer that has the ability to print in slices as small as 16 microns \cite{printer}. A printer with such fine resolution is necessary to capture the minute details of the mapped fingerprint onto $M$. As in \cite{3Dfingers}, the mold is printed in 30 micron layers as this captures the necessary detail of the mapped fingerprints, while simultaneously decreasing the  print time of $M$ from 8 hours to 4 hours \cite{3Dfingers}. At the conclusion of printing, the mold is soaked in 2M NaOH\footnote{NaOH (Sodium Hydroxide) is a basic (alkaline) solution that cleans the residual printing support material away from the mold.} for about 4 hours to dissolve away the support material from the printed mold in a manner that does not damage the fingerprint ridges. After chemical cleaning a high fidelity fingerprint mold is ready for casting fingerprint targets (Fig. 6).

The resultant mold will only produce a solid casting, since casting material will fill the entire mold cavity. To make the cast wearable (e.g. mounting to a robotic gripper) or manual evaluation (e.g. human placement of the target) a {\it``scaffolding framework"} $F$ is fabricated, which, when used in conjunction with $M$, creates a wearable 3D target $T$ (Fig. 7). The process for generating $F$ is further expounded upon below.

\begin{figure}[h]
\begin{center}
\includegraphics[scale=.2]{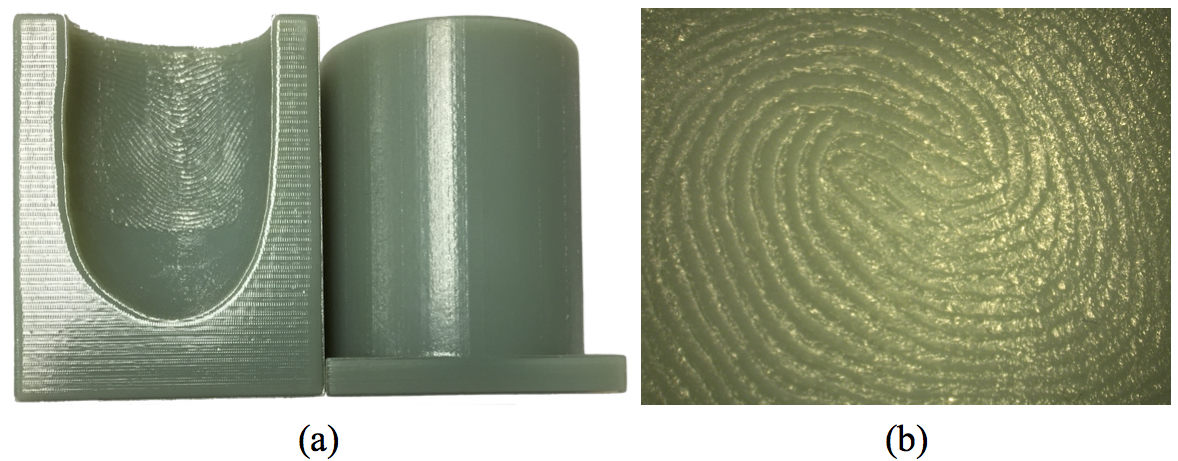}
\caption{(a) High fidelity 3D printed fingerprint mold $M$. (b) View of fingerprint engraving on $M$ at 20X magnification. The magnified image in (b) shows that all the friction ridge patterns are clearly present in the mold M. These friction ridge patterns are inverted, since negative molds are necessary to produce positive fingerprint targets (Fig 7 (c)).}
\end{center}
\vspace{-1.5em}
\end{figure} 

\begin{figure}[h]
\begin{center}
\includegraphics[scale=.15]{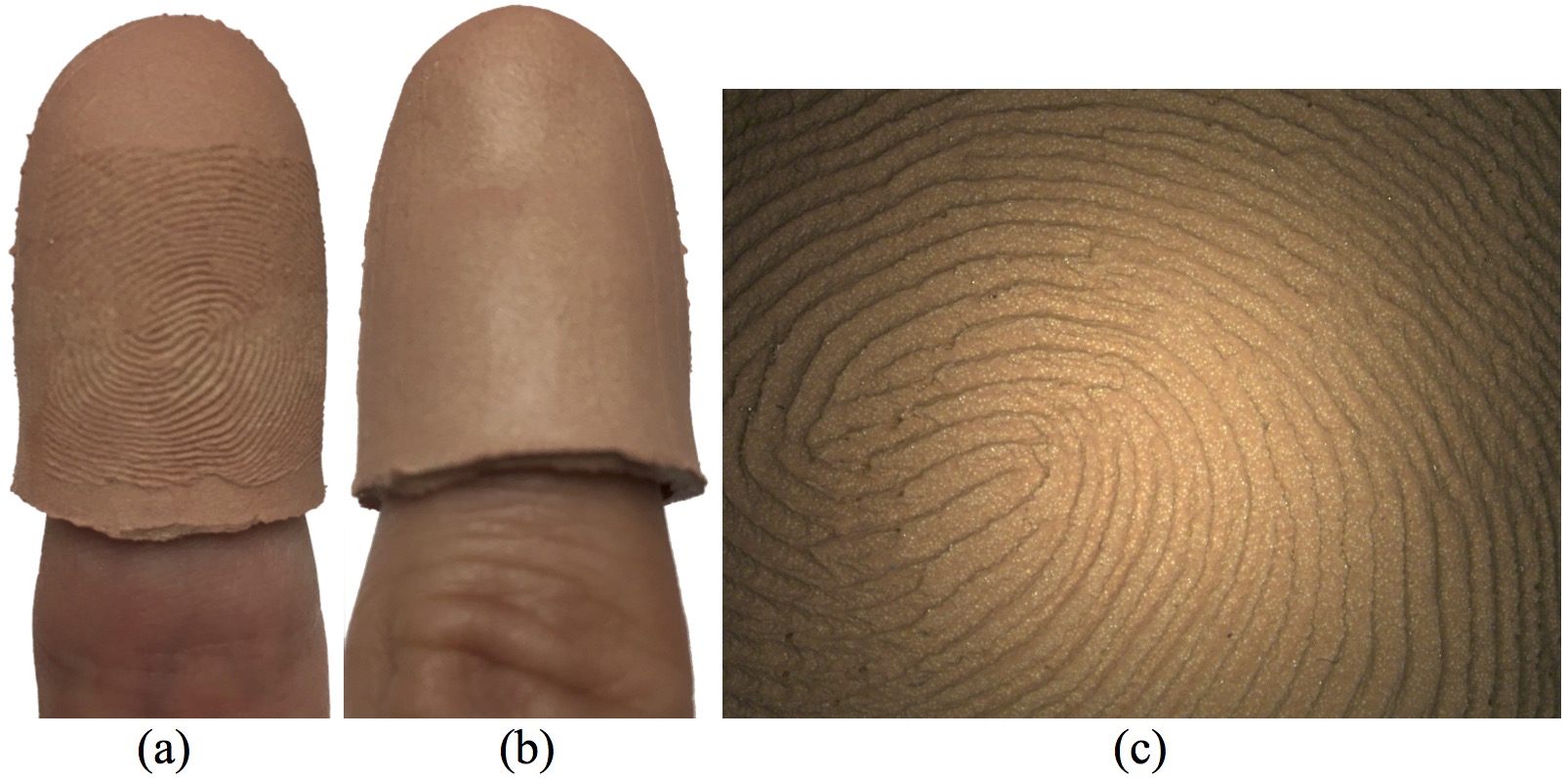}
\caption{3D wearable Universal Fingerprint Target (a) front view, (b) rear view, and (c) view of the Universal Fingerprint Target ridges at 20X magnification.}
\end{center}
\vspace{-1.5em}
\end{figure}

\begin{figure*}[t]
\begin{center}
\includegraphics[scale=.25]{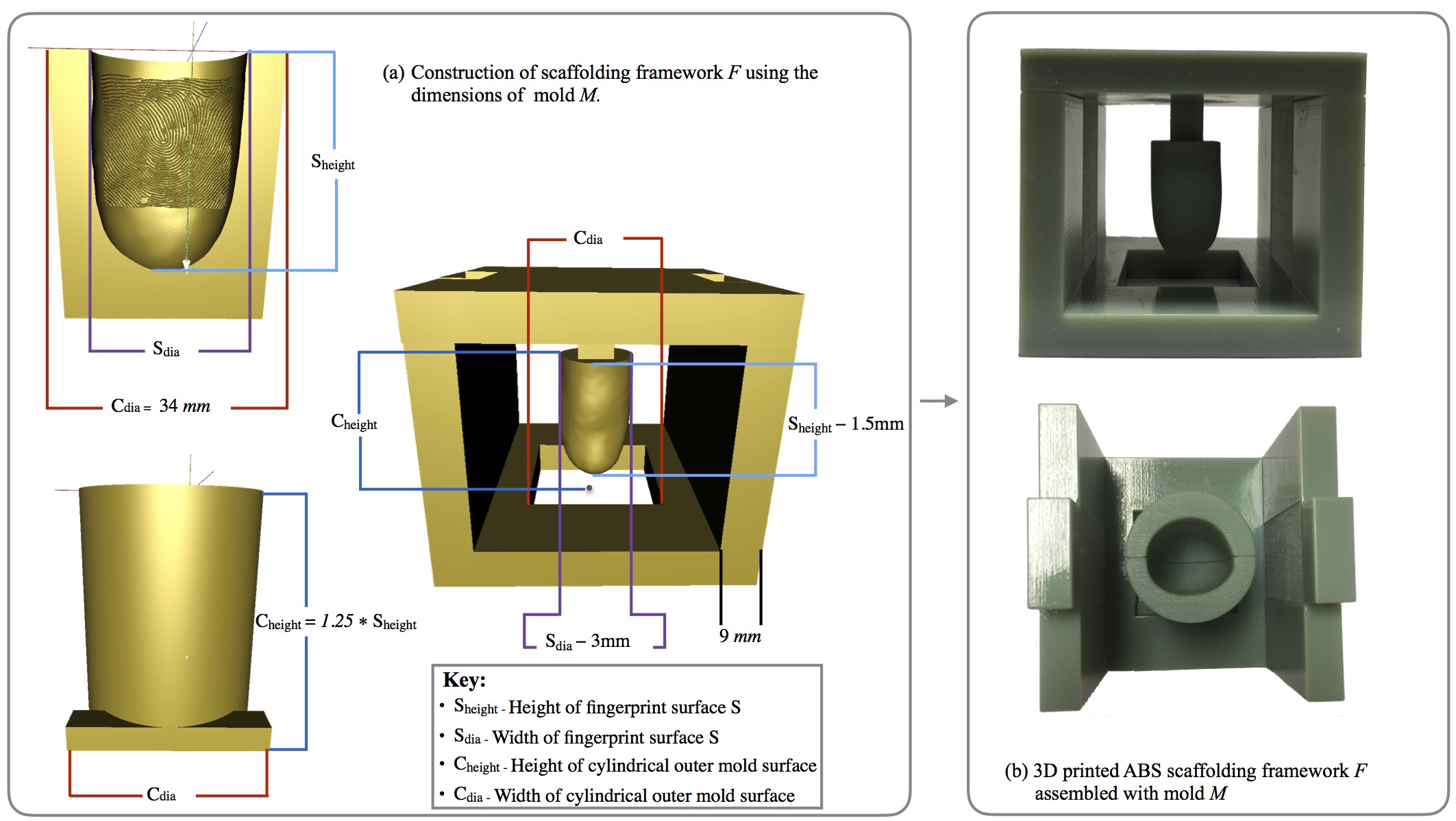}
\caption{Fabricating scaffolding $F$ using the dimensions of the mold, $M$. (a) scaffolding framework $F$ is electronically modeled; (b) the electronic scaffolding system is physically generated in acrylonitrile butadiene styrene (ABS) using a high resolution 3D printer. Using $F$ in conjunction with $M$, 3D wearable fingerprint targets $T$ are repeatably produced.}
\end{center}
\vspace{-1.5em}
\end{figure*} 

\subsection{Scaffolding Fabrication}

To create a wearable fingerprint casting, a hollow, appropriately shaped void must be cured into the casted material as it resides in $M$. This void enables wearability as it creates the space where an end user's finger (or robotic attachment) would reside during evaluation.

We build upon the above idea by developing (based upon the dimensions $M$) a scaffolding framework $F$ used to insert a fingerprint surface $S'$ (with diameter slightly smaller than $S_{dia}$) into $M$ during successive fingerprint target casts (Fig. 8 (a)). In doing so, we ensure that when casting material is injected into the mold, the space between $S$ and $S'$ will be filled to form a wearable fingerprint target $T$.

The scaffolding $F$ consists of several components: a base platform that holds the mold $M$ in place, two sides extending beyond the top of $M$, and a top piece from which the fingerprint surface $S'$ is suspended. Aside from $S'$, all of these pieces are generated by creating a simple cuboid shape and applying affine transformations until the component is of the correct size and in the correct position. The thickness of scaffolding walls is chosen to be 9 mm, which provides the structural robustness and durability needed for repeated castings of fingerprint targets. In addition, a concentric rectangular prism is cut from the inside of the base component. The length and width of this rectangular prism share the same dimension ($C_{dia}$) as the diameter of $M$. This ensures that $M$ will attach securely into the base unit, thus controlling the thickness of the casted targets. Since the diameter of $M$ is fixed (based upon the 95\textsuperscript{th} percentile of the human finger width at 34 mm), any mold can be attached interchangeably into a single scaffolding system.

Given that $S'$ is a fingerprint surface with a diameter smaller than $S$, we can derive $S'$ from the same scanned fingerprint surface that we originally used to generate $S$. That is, given a smooth scanned 3D fingerprint surface $S_{smooth}$, we can generate $S'$ by shrinking $S_{smooth}$ along the direction of its normals by 1.5 mm. More formally, if $v_1 = \begin{bmatrix} v_x & v_y & v_z \end{bmatrix}$ is a vertex of $S_{smooth}$ and $n_1 = \begin{bmatrix} n_x & n_y & n_z \end{bmatrix}$ is the corresponding normal vector to $v_1$, then generating the new vertex $v' = \begin{bmatrix} v'_x & v'_y & v'_z \end{bmatrix}$ for $S'$ is computed as:
\begin{equation}
v' = \begin{bmatrix} v_x \\ v_y \\ v_z \end{bmatrix} - \begin{bmatrix} n_x \\ n_y \\ n_z \end{bmatrix} \times 1.5
\end{equation}
After all vertices of $S_{smooth}$ have been iteratively shrunken along the direction of their corresponding normals, the top of $S'$ is stitched shut using a triangle fan\footnote{A triangle fan is a circular mesh surface, formed by placing a center vertex and filling in the circle with triangles that all share the center vertex.}. 

As with $M$, the electronic model of $F$ is 3D printed using the same high resolution printer and parameters (Fig 8 (b)). $F$ is also cleaned with 2M NaOH solution to remove residual printing support material. Although $F$ does not have the minute detail that $M$ does, high resolution printing is still needed for printing $F$ so that registration between $F$ and $M$ is consistent and reproducible. This ensures the high fidelity of the casted targets is preserved.

Upon completed fabrication of both $M$ and $F$, we now have tools for repeatably generating high fidelity, 3D wearable fingerprint targets $T$. In the following sections, we discuss and illustrate what material properties are required in fabricating targets, and what casting techniques should be followed in order to preserve the fidelity of the final 3D wearable fingerprint targets.

\section{Casting Material}
In this section, we discuss the characteristics necessary (to emulate human skin) in the casting material for the 3D Universal Fingerprint Target. Additionally, we prescribe a process for concocting a material consisting of these characteristics.

\subsection{Material Characteristics} Our material selection needs to carefully consider the optical, electrical, and mechanical properties inherent to the human finger. 
\begin{itemize}
\item {\bf Optical Property:} Optical readers rely on proper reflectance and refraction of light rays on the human finger surface to detect a fingerprint. Therefore, the optical properties of the targets must be the same as that of human skin to be accurately sensed by optical readers. Materials that are black will improperly absorb all light rays and materials of high reflectivity will improperly scatter all light rays, in both cases preventing targets of these materials from being imaged by many optical readers. 
\item {\bf Electrical Property:} In addition to the color attribute, the targets must also be inherently conductive to act as a conductive plate and create capacitive differences between ridges and valleys on the cells within the semiconductor chips on capacitive sensors. 
\item {\bf Mechanical Property:} Finally, the mechanical properties of the target material must lie within the range inherent to the human epidermis to ensure high quality fingerprint target image acquisition. Materials that deviate from the elasticity of the human epidermis could negatively impact the target in several ways. If the elasticity is too large, the minute details of the minutia will be lost as the target is compressed against the sensor and the ridges collapse under the force being exerted (Fig. 9 (a)). If, on the other hand, the elasticity is too small, or the hardness is too great, the fingerprint target will not flatten around the sensor platen, resulting in only partial print images of the fingerprint surface (Fig. 9 (b)).
\end{itemize}

\begin{figure}[h]
  \centering
  \subfloat[900 \% elongation at break]{\includegraphics[scale=2.25]{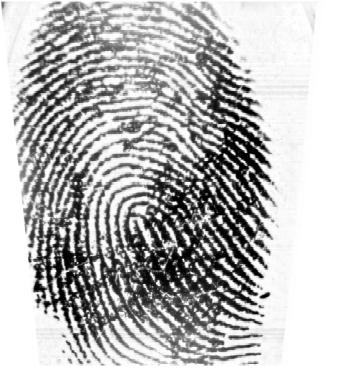}\label{fig:f1}}
  \hfill
  \subfloat[Shore A 50]{\includegraphics[scale=2.25]{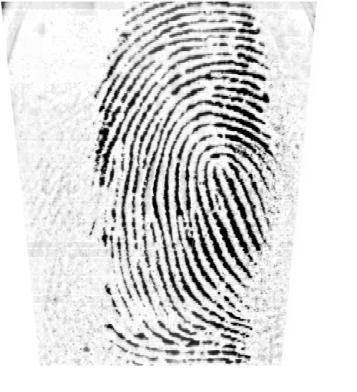}\label{fig:f2}}
  \caption{Fingerprint impressions captured from targets lacking proper mechanical characteristics. Notice (a) the presence of aberrations resulting from excessive elasticity in the target and (b) partial impression due to excessive hardness of the target.}
  \vspace{-1.5em}
\end{figure}

\begin{figure*}[t]
\begin{center}
\includegraphics[scale=.3]{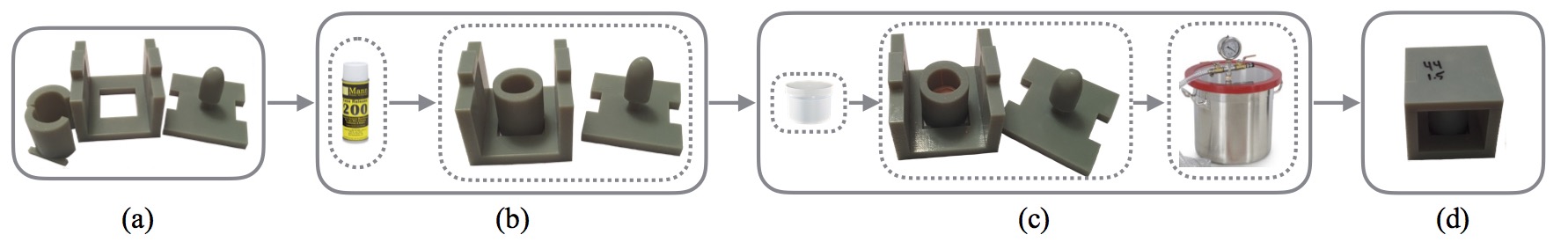}
\caption{Casting Process Flow. (a) The unassembled mold and scaffold; (b) the mold has been sprayed with release agent and clipped into the scaffolding base; (c) the material is poured into the mold and vacuum degassed; (d) the top component is clipped onto the scaffold in order to make the casted targets wearable. The mold is left to cure for 72 hours.}
\end{center}
\vspace{-1.5em}
\end{figure*}

\subsection{Material Fabrication} 

To achieve the optical, capacitive, and mechanical criteria necessary for the universal fingerprint target, several off-the-shelf materials are mixed to form a single casting mixture which encapsulates all the desired material characteristics. 

The bulk of the mixed material consists of conductive silicone (SS-27S) \cite{ss-27s}, which is a mixture of conductive particles (silver coated aluminum) into a base polymer (polydimethylsiloxane (PDMS)) \cite{ss-27s} at a percolation threshold\footnote{The percolation threshold is the point at which enough conductive particles have been introduced into a base polymer, such that electrons can flow through the polymer \cite{chemistry}.}. Although polymers can be made very conductive at the percolation threshold, unfortunately, the mechanical properties of the material are altered. In particular, the silicone becomes harder, less elastic, and thixotropic\footnote{Thixotropic materials are very difficult to pour into a mold, since such materials resist flowing without external force being exerted on them.}. Therefore, in order to bring SS-27S back down in hardness (to the level of the human finger) and make the material less thixotropic (less viscous), a silicone thinner is introduced into the mixture.

By mixing 4 \% silicone thinner into SS-27S, the Shore A durometer value is reduced from 50 to 38.5 and the viscosity is reduced from 50,000 cps to 30,250 cps. Using SS-27S in conjunction with silicone thinner provides an electrically conductive material with similar mechanical properties to the human skin. Furthermore, the decreased viscosity of the uncured material (caused by the thinner), enables easier casting.

The final component still missing from the casting material is optical similarity to the human skin. To introduce the optical characteristic into the casting material, a flesh-toned pigment (Pantone PMS 488C) is mixed into the SS-27S and thinner mixture \cite{pigments}\cite{pantone}. In accordance with the technical data sheet for the pigment, 3 \% pigment (by mass) is added to the casting material.

With all three of the aforementioned components in place (SS-27S, thinner, pigment), vigorous mechanical sheer mixing is performed to properly and uniformly mix the components. A dual asymmetric centrifugal (DAC) sheer mixing machine is used to perform this uniform mixing \cite{mixer}. In addition to uniform mixing, this machine prevents air from entering into the casting mixture. In our experiments, 60 grams of SS-27S, 2.4 grams of thinner, and 1.8 grams of pigment were mixed for 30 seconds in the sheer mixer at 1500 rpm. At the conclusion of the DAC mixing, a casting material containing all the characteristics\footnote{A simpler casting material - useful for interoperability assessment of contact and contactless optical readers - can be fabricated by mixing (with the FlakTek) pure PDMS and PMS 488C pigment. These targets are not conductive, and are therefore unusable for capacitive reader evaluation, but they are optically and mechanically similar to the human finger and are cheaper to manufacture (Fig. 2 (d)) (Table 1).} necessary for fingerprint targets to register on the most commonly used fingerprint readers is ready for casting.

\section{Casting Process}

To properly and repeatedly cast high fidelity, 3D, wearable fingerprint targets, the following casting process using our mold, scaffolding framework, and casting material is prescribed. 

{\it (i)} To facilitate a clean extraction of the fingerprint targets from the mold and scaffolding framework, both the mold, and the inner finger surface are aerosol spray coated with silicone release agent \cite{release}. After the silicone release agent has been applied and given fifteen minutes to dry, the two mold components are attached securely into the scaffolding base (Fig. 10 (b)). 

{\it (ii)} 8 grams of the casting material are transferred from a mixing container to the mold - via a disposable pipet (Fig. 10 (c)). It was experimentally determined that 8 grams of casting material is sufficient to fill the mold cavity. To remove air bubbles introduced during the mechanical transfer of material, the entire scaffolding framework (with material inside) is placed into a vacuum degaser at 98 kPa (0.97 atm). At this pressure, all of the air bubbles introduced (during the mechanical transfer of material) are removed. 

{\it (iii)} Finally, the top component of the scaffolding framework is inserted into the casting material, and attached securely to the scaffolding base (Fig. 10 (d)). By inserting this top component, wearability is added to the final fingerprint castings. With the scaffolding system fully assembled, high fidelity, 3D, wearable, universal fingerprint targets can be carefully extracted from the mold and scaffolding framework after 72 hours (Fig. 11). 

This casting process can be repeated to manufacture hundreds\footnote{This is a coarse estimate based on the reported toughness of the digital ABS mold material \cite{abs}.} of fingerprint targets from a single mold and scaffolding system.

\begin{figure}[h]
\vspace{-2.5em}
  \centering
  \subfloat[]{\includegraphics[scale=.22]{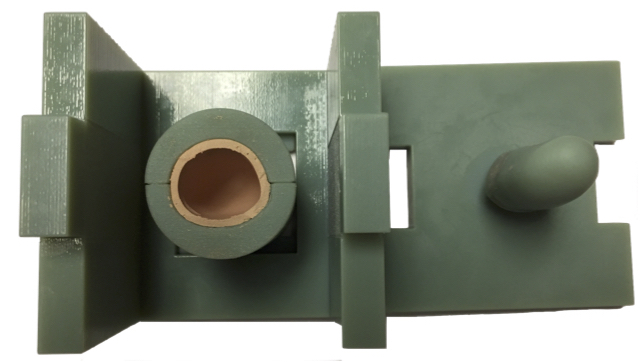}\label{fig:f1}}
  \hfill
  \subfloat[]{\includegraphics[scale=.22]{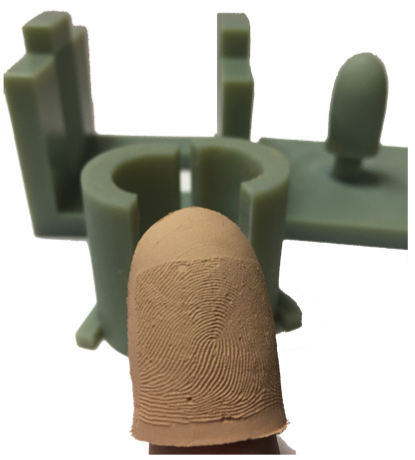}\label{fig:f2}}
  \caption{Extraction of high fidelity Universal 3D Wearable Fingerprint Target. (a) Extract top component from casting; (b) Remove wearable fingerprint target from mold}
\vspace{-1.0em}
\end{figure}

\section{Target Fidelity and Reproducibility}

To establish the universal fingerprint targets as standard evaluation artifacts, we must show that the proposed fabrication process (i) is of high fidelity and (ii) is reproducible. Both of these criterion are verified in the following subsections.

\begin{figure*}[h]
\begin{center}
\includegraphics[scale=.25]{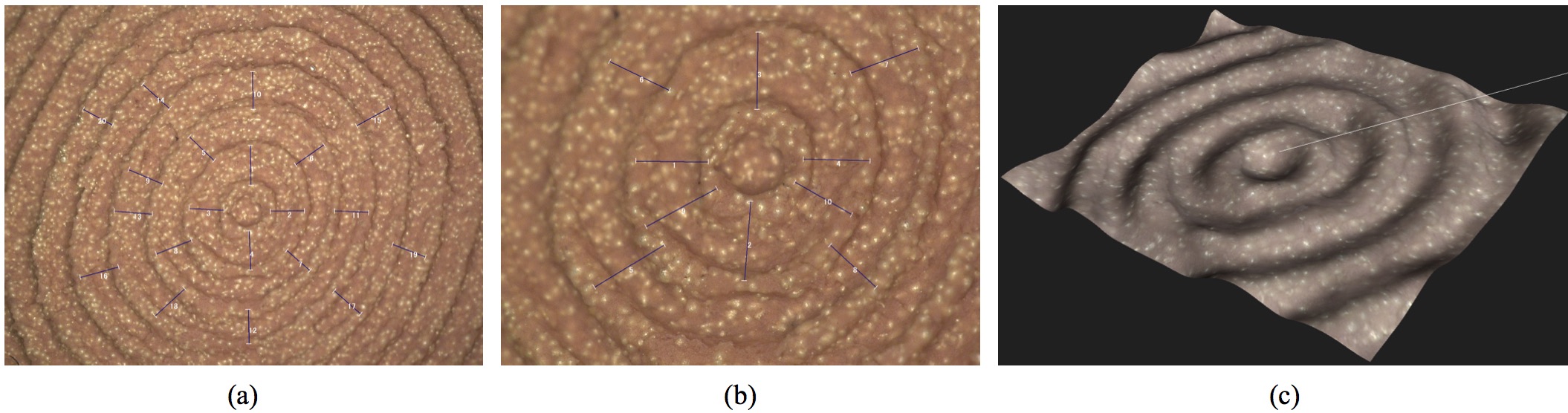}
\caption{Images of the universal fingerprint target (mapped with circular sine gratings) captured using a Keyence optical microscope \cite{microscope}. Point-to-point ridge distances are measured. (a) Image at 50X magnification and annotated with 20 point-to-point distances. (b) Image at 100X magnification and annotated with 10 point-to-point distances. (c) 3-D image generated by the microscope which qualitatively illustrates the uniformity in ridge height of the circular gratings on the universal fingerprint target. The granular texture in (a), (b), and (c) is evidence of the aluminum coated silver particles mixed into the universal fingerprint target which allows the target to be imaged by capacitive fingerprint readers.}
\end{center}
\vspace{-1.25em}
\end{figure*}

\subsection{Fidelity}

A 3D universal fingerprint target is of high fidelity if its 3D ridges retain the topology inherent to the original 2D image it was fabricated from. We posit that fidelity of universal fingerprint targets can be objectively determined and compensated by quantifying the errors (as a deviation of the 3D target topology from the topology of the 2D mapping pattern) at each step in the fabrication process (Fig. 4) and accounting for these errors during fabrication. 

{\it (i) Error in Electronic Modeling of Fingerprint Mold -} Arora et al. \cite{3Dfingers} showed that the projection algorithm used to map the 2D fingerprint image to a 3D finger surface results in a 5.8 \% decrease in point-to-point distances inherent to the original 2D fingerprint image. Because the electronic fabrication of the fingerprint mold (Fig. 4 (a)) uses the same 2D to 3D projection algorithm as \cite{3Dfingers}, the same error will be encountered in our universal fingerprint target fabrication process.

{\it (ii) Error in 3D printing -} Arora et al. \cite{3Dfingers} also observed an 11.42 \% decrease in point-to-point distances (inherent to the original 2D fingerprint image) when fabricating the physical 3D target on a high resolution 3D printer. Since printing the fingerprint mold in (Fig. 4 (b)) was performed using the same printer as \cite{3Dfingers}, the universal fingerprint target fabrication process will encounter the same error.

In total, the electronic mold fabrication in (Fig. 4 (a)), and the physical realization of that electronic model in (Fig. 4 (b)) result in a 17.22 \% decrease in point-to-point distances on the printed mold in comparison to the original 2D image. While this error may seem significant, it can be rectified (as shown in \cite{goldfinger}) by setting the scale during 2D/3D projection from 19.685 pixels/mm to 16.79 pixels/mm. In doing so, the errors introduced during mold modeling (Fig. 4 (a)) and 3D mold printing (Fig. 4 (b)) are compensated.

{\it (iii) Error in Casting -} The fidelity in the universal fingerprint target post casting (Figs. 4 (d), (e)) is validated in the following manner. First, three universal fingerprint target castings are fabricated using three different molds; each mapped with different 2D calibration patterns (vertical, horizontal, and circular sine gratings with a frequency of 10 pixels). At a projection scale of 16.79 pixels/mm (at 500 ppi) and 17.22 \% reduction in point-to-point distances during electronic modeling and 3D printing, 10 pixel ridge distances on the calibration pattern should correspond to an actual ridge distance of 0.508 mm on the casted calibration target. Using an optical microscope, 5 images of each universal fingerprint target are captured at both 50X magnification and 100X magnification (Fig. 12) \cite{microscope}. A software tool available with the optical microscope is used to mark 20 point-to-point ridge distances at 50X magnification and 10 point-to-point ridge distances at 100X magnification in all the acquired optical microscope images. The microscope software was calibrated using a micrometer resolution calibration target. Table 2 shows the average point-to-point ridge distances at both magnifications for all 3 casted targets. In comparison to the ground truth distance of 0.508 mm, the optical microscope reveals the empirical mean point-to-point ridge distances to be 0.499 mm, attributing to a 1.8 \% reduction in point-to-point distances on the universal fingerprint target during casting. This reduction of 1.8 \% in point-to-point ridge distances is not unexpected, since the conductive silicone used to fabricate the universal fingerprint targets is estimated to shrink by 2 \% during vulcanization. Again, this error can be compensated by adjusting the projection scale during 2D/3D mapping. 

\begin{table}[t]
 \centering
 \caption{Average point-to-point ridge distances observed on universal fingerprint targets, measured using the Keyence Optical Microscope at 50X and 100X magnification. The expected point-to-point ridge distance is 0.508 mm. (standard deviation is recorded in parenthesis).}
\begin{tabular}{ |c||c|c|}
 \hline
 Calibration Pattern & 50X Magnification & 100X Magnification \\
 \hline
 \hline
 Vertical Gratings & 0.509 mm (.031) & 0.496 mm (.023) \\
 \hline
 Horizontal Gratings & 0.501 mm (.026) & 0.490 mm (.028)\\
  \hline
 Circular Gratings & 0.513 mm (.029) & 0.486 mm (.035)\\
  \hline
\end{tabular}
\vspace{-1.5em}
\end{table}

{\it (iv) End-to-end Error} In this final error analysis, the full, end-to-end fabrication process is scrutinized. More specifically, an experiment is conducted which demonstrates that features present on a 2D fingerprint image are preserved after converting the 2D fingerprint image into a wearable, 3D, universal fingerprint target. 

To conduct this experiment, six different universal fingerprint target molds are fabricated using six fingerprint images from the NIST SD4 database \cite{sd4}. Subsequently,  six universal fingerprint targets are cast from the fingerprint molds. Finally, comparison scores are generated between the NIST SD4 rolled fingerprint images and 2D fingerprint images acquired from the corresponding six universal fingerprint targets. Fingerprint images of the universal fingerprint targets are obtained using an Appendix F certified, 500 ppi, optical reader. Figure 13 illustrates corresponding minutia points between a NIST SD4 rolled fingerprint image and a fingerprint image acquired from its corresponding universal fingerprint target. Table 3 reports similarity scores for each of the six universal fingerprint targets in comparison to the NIST SD4 rolled print used to fabricate them.

\begin{figure}[!t]
\begin{center}
\includegraphics[scale=.135]{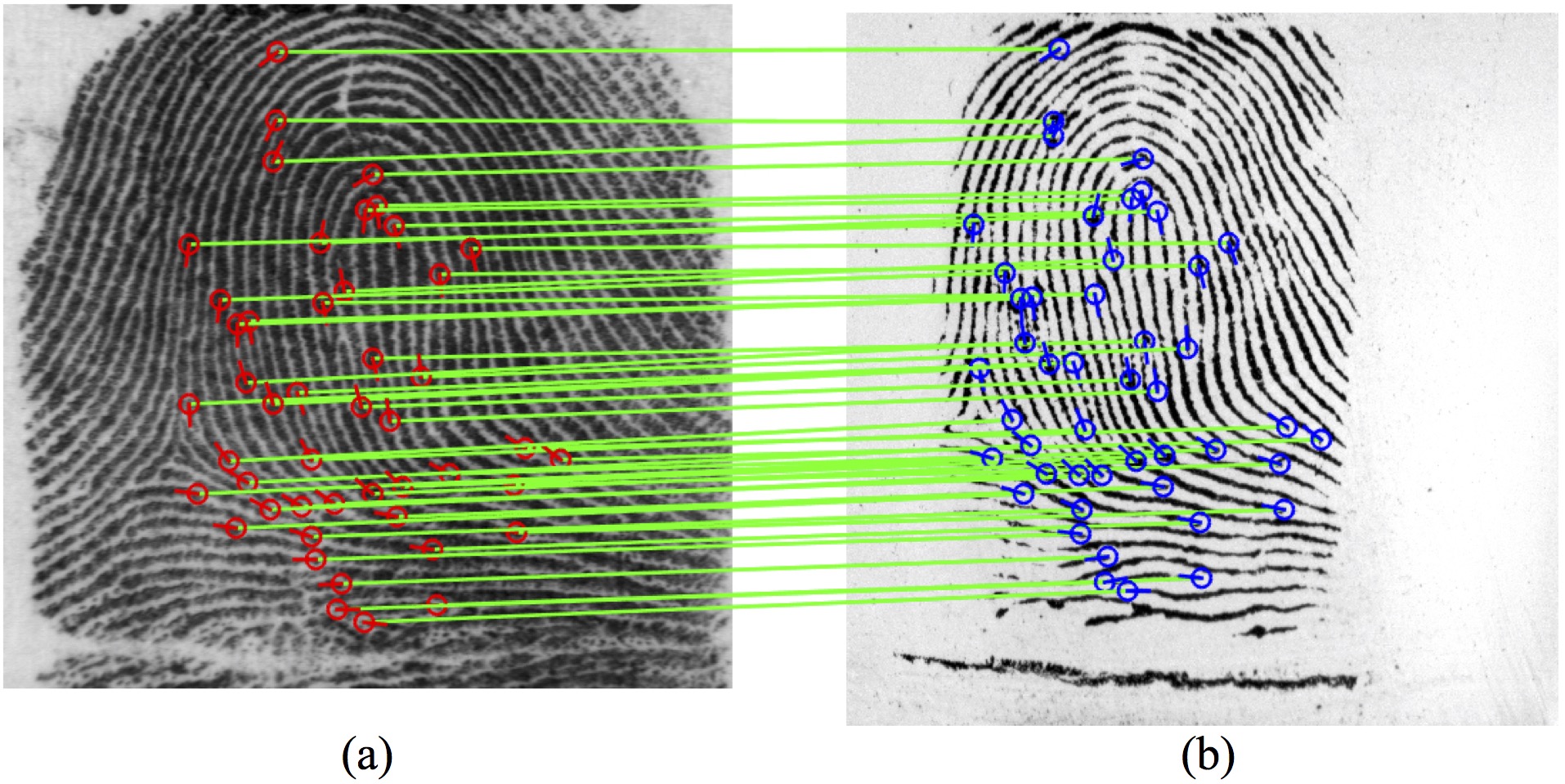}
\caption{Comparing the source fingerprint image to the image of the corresponding universal fingerprint target. (a) NIST SD4 S0083 rolled fingerprint image is compared to (b) a universal fingerprint target image; (b) is fabricated using (a) and imaged using an Appendix F certified, optical, 500 ppi fingerprint reader. A similarity score of 608 is computed between (a) and (b) using Verifinger 6.3 SDK (threshold is 33 at FAR=0.01 \%). The minutia points in correspondence between (a) and (b) are shown.  }
\end{center}
\vspace{-1.5em}
\end{figure} 

The key findings of this experiment are as follows:

\begin{itemize}

\item The corresponding minutia points between images captured using the universal fingerprint targets and the images used to generate each target (Fig. 13) show that salient 2D features inherent to the NIST rolled fingerprint images are retained following their fabrication into a universal fingerprint target.

\item The universal fingerprint targets (Table 3) almost always outperform previous 3D optical targets \cite{3Dfingers} (Table 4) by achieving higher similarity scores between the finished 3D target images and the ground truth image used to fabricate the respective target. Furthermore, the universal fingerprint targets perform comparably to goldfingers \cite{goldfinger} on capacitive readers (Tables 3 and 4).

\item Unlike past research in 3D fingerprint targets, the universal fingerprint target achieves comparison scores on contactless-optical readers well above the acceptance threshold of 33. We do note that the universal fingerprint targets achieve lower comparison scores against the SD4 images when using the contactless-optical reader as opposed to the contact-optical reader for image acquisition. One plausible explanation is that the universal fingerprint targets have a ridge height greater than the ridge height of the adult human finger\footnote{The ridge height of the universal fingerprint targets is set to 0.33 mm instead of the 0.06 mm ridge height of the adult human finger. This is due to limitations in the printing resolution of current state of the art 3D printers. In the future, we will explore novel techniques for fabricating the mold which enable even higher resolution than 3D printing.}. This discrepancy may cause errors as the contactless-reader unrolls a 3D fingerprint into a 2D fingerprint image.  

\end{itemize}

\begin{table}[t]
 \centering
  \begin{threeparttable}
 \caption{Universal Fingerprint Target Similarity Scores\tnote{1} (SD4 fingerprint image vs. corresponding target image). Proposed Targets.}
\begin{tabular}{ |c||c|c|c|}
 \hline
 \specialcell{SD4 \\Fingerprint} & \specialcell{Contact \\Optical Reader \\(500 ppi)} & \specialcell{Contactless \\Optical Reader \\ (500 ppi)} & \specialcell{Capacitive \\ Reader \\(500 ppi)} \\
 \hline
 \hline
 S0005 & 584 & 152 & 161\\
 \hline
 S0010 & 539 & 137 & 305\\
  \hline
 S0031 & 600 & 105 & 221\\
  \hline
 S0044 & 498 & 150 & 323\\
  \hline
 S0068 & 327 & 146 & 368\\
  \hline
 S0083 & 608 & 176 & 323\\
  \hline
\end{tabular}
\begin{tablenotes}
\item[1] Verifinger 6.3 SDK was used for generating similarity scores. The score threshold at 0.01 \% FAR is 33. Verifinger was chosen so that comparisons could be made between the universal fingerprint targets and previous studies \cite{3Dfingers}\cite{whole_hand}\cite{goldfinger}
\end{tablenotes}
\end{threeparttable}
\end{table}

\begin{table}[t]
 \centering
 \begin{threeparttable}
\caption{3D Printed Target \tnote{1} Similarity Scores (SD4 fingerprint image vs. corresponding target image). Targets from \cite{3Dfingers, goldfinger, whole_hand}.}
\begin{tabular}{|c||c|c|}
 \hline
 \specialcell{SD4 Fingerprint} & \specialcell{Contact-Optical Reader \\(500 ppi) \tnote{2}} & \specialcell{Capacitive Reader\\(500 ppi) \tnote{3}}\\
 \hline
 \hline
 S0005 & 719 & 471\\
 \hline
 S0010 & 129 & 333\\
  \hline
 S0031 & N/A & N/A\\
 \hline
  S0044 & 371 & N/A\\
  \hline
 S0068 & N/A & N/A\\
 \hline
 S0083 & 441 & 183\\
\hline
\end{tabular}
\begin{tablenotes}
\item[1] These targets were fabricated using processes reported in \cite{3Dfingers, goldfinger, whole_hand}. They are not interoperable across optical and capacitive readers as are the Universal Fingerprint Targets.
\item[2] Best results taken from \cite{3Dfingers} \& \cite{whole_hand}
\item[3] Results taken from \cite{goldfinger}
\end{tablenotes}
\end{threeparttable}
\vspace{-.5 em}
\end{table}

In summary, the 2D ground truth fingerprint features are found to be preserved during fabrication into a 3D universal fingerprint target and subsequent image acquisition (with high accuracy) by contact-based optical readers, contactless-optical readers, and capacitive readers. In other words, the universal fingerprint target is highly interoperable across different fingerprint reader technologies commonly in use.

\subsection{Reproducibility}

In the previous section, the fabrication process for creating universal fingerprint targets was quantitatively shown to be of high fidelity. One remaining criterion that must be objectively verified to solidify the use of universal fingerprint targets as standardized evaluation artifacts is the reproducibility of high fidelity universal fingerprint target fabrication. To that end, we individually examine the reproducibility of each step in the universal fingerprint target fabrication process. 

The electronic model of the universal fingerprint target mold and scaffolding system can be easily reproduced by simply executing a program. Additionally, the mold and scaffolding system can be physically reproduced via 3D printing with accuracy as high as 20 microns \cite{printer}. Therefore, the only step in the universal fingerprint target fabrication process that must still be verified as reproducible is the casting step.

To demonstrate reproducibility in casting, 12 universal fingerprint targets are fabricated from 6 fingerprint molds. The 12 universal fingerprint targets correspond to 6 different targets each fabricated 2 times (with a time lapse of several weeks between target replication). Each mold is mapped with one of 6 NIST SD4 rolled fingerprint images (S0005, S0010, S0031, S0044, S0068, and S0083). Let the two sets of universal fingerprint targets be formally defined as $T_1$ and $T_2$, where $T_1$ is the first set of castings and $T_2$ is the set of castings produced several weeks later.
 
  \begin{table}[t]
 \centering
 \begin{threeparttable}
\caption{Specifications of the Fingerprint Readers Used in Our Experiments}
\begin{tabular}{|c||c|c|c|c|}
 \hline
 \specialcell{Reader \\NDA Alias\tnote{1}} & \specialcell{Reader Type} & \specialcell{Resolution} & \specialcell{Certifications}\\
 \hline
 \hline
 $COR\_A$  & \specialcell{Contact-Optical} & 500 {\it ppi} & Appendix F \\
 \hline
 $COR\_B$  & \specialcell{Contact-Optical} & 500 {\it ppi} & Appendix F\\
  \hline
 $CLOR$  & \specialcell{Contactless-Optical} & 500 {\it ppi} & PIV \\
  \hline
 $CPR\_A$  & \specialcell{Capacitive} & 500 {\it ppi} & PIV\\
 \hline
 $CPR\_B$ & \specialcell{Capacitive} & 500 {\it ppi} & PIV\\
 \hline
\end{tabular}
\begin{tablenotes}
\item[1] Because of a Nondisclosure agreement (NDA) with our vendors, we do not release the names of the fingerprint readers.
\end{tablenotes}
\end{threeparttable}
\vspace{-1.5 em}
\end{table}

 \begin{table*}[t]
 \centering
 \begin{threeparttable}
\caption{Universal Fingerprint Target Genuine Similarity Scores\tnote{1} (SD4 Fingerprint Image vs. Corresponding Target Image) Mean and (Standard Deviation) of Scores for 10 Impressions are Reported }
\begin{tabular}{|c||c||c|c|c|c|c|c|}
 \hline
 \specialcell{Target Set} & \specialcell{Reader} & \specialcell{S0005} & \specialcell{S0010} & \specialcell{S0031} & \specialcell{S0044} & \specialcell{S0068} & \specialcell{S0083}\\
 \hline
 \hline
 $T_1$  & \specialcell{Contact-Optical ({\it COR\_A})} & 212.2 (19.97) & 200.3 (15.76) & 204.4 (19.28) & 177.0 (23.13) & 141.3 (8.99) & 254.5 (20.09) \\
 \hline
 $T_2$  & \specialcell{Contact-Optical ({\it COR\_A})} & 247.4 (10.32) & 207.0 (9.17) & 226.8 (17.31) & 230.6 (16.04) & 166.5 (10.01) & 248.7 (8.44)\\
  \hline
  \hline
 $T_1$  & \specialcell{Contactless-Optical ({\it CLOR})} & 203.9 (15.75) & 127.5 (15.22) & 140.6 (7.83) & 154.3 (11.76) & 169.9 (17.29) & 172.8 (17.81)\\
  \hline
 $T_2$  & \specialcell{Contactless-Optical ({\it CLOR})} & 205.1 (14.09) & 134.3 (23.01) & 150.7 (10.28) & 143.4 (15.73) & 170.5 (22.16) & 172.0 (11.51)\\
 \hline
 \hline
 $T_1$ & \specialcell{Capacitive ({\it CPR\_A})} & 163.2 (21.81) & 128.6 (25.22) & 177.1 (14.47) & 141.3 (22.74) & 121.3 (14.20) & 190.9 (17.14)\\
 \hline
$T_2$ & \specialcell{Capacitive ({\it CPR\_A})} & 188.1 (16.68) & 183.8 (16.50) & 194.4 (21.63) & 173.3 (10.42) & 156.5 (6.95) & 194.0 (6.65)\\
 \hline
\end{tabular}
\begin{tablenotes}
\item[1] Innovatrics matcher was used to generate similarity scores. The threshold of the matcher at FAR = 0.01 \% was computed to be 49 on the FVC 2002 and 2004 databases \cite{fvc2002, fvc2004}.
\end{tablenotes}
\end{threeparttable}
\vspace{-.5 em}
\end{table*}
 
Next, the average and standard deviation of genuine scores between 10 impressions from each target in the two target sets $T_1$ and $T_2$ collected on 3 types of fingerprint readers ({\it COR\_A}, {\it CLOR}, and {\it CPR\_A} (Table 5)) and the corresponding fingerprint image in NIST SD4 are computed using the Innovatrics fingerprint SDK\footnote{We use the Innovatrics fingerprint SDK since we recently acquired this matcher, and it is shown to have high accuracy. Mention of any products or manufacturers does not imply endorsement by the authors or their institutions of these products or their manufacturers.} \cite{in}. The averages and standard deviations of genuine similarity scores between target impressions from each target in $T_1$ and its corresponding fingerprint image in NIST SD4 are formally defined as $GS_1$. Conversely, $GS_2$ is defined as the averages and standard deviations of genuine similarity scores between target impressions for each target in $T_2$ and its corresponding fingerprint image in NIST SD4.

By analyzing the means of the similarity scores in $GS_1$ and $GS_2$, reproducibility in casting universal fingerprint targets is verified. In particular, by showing that the means of the similarity scores in $GS_1$ and $GS_2$ are all well above the genuine acceptance threshold, we demonstrate that targets (from multiple castings) in $T_1$ and $T_2$ are all of high fidelity, since impressions from both sets of targets (on multiple types of fingerprint readers) achieve high similarity scores against the ground truth images (SD4) from which they were fabricated. The means and standard deviations of the genuine similarity scores in $GS_1$ and $GS_2$ are reported in Table 6. 

We note that the means of all similarity scores in $GS_1$ and $GS_2$ are within 0.72 \% when using the contactless fingerprint reader for image acquisition (Table 6). This indicates high similarity between 3D fingerprint topologies on targets in $T_1$ and $T_2$. Additionally we note that the means of similarity scores in $GS_1$ and $GS_2$ differ slightly when using contact based fingerprint readers for image acquisition. This is not surprising since the targets in $T_1$ were fabricated with smaller amounts of silicone thinner than the targets in $T_2$. As such, the softer targets in $T_2$ morphed around the fingerprint reader platen more than the targets in $T_1$ and produced images with larger friction ridge area and number of minutia (recall Fig. 9 (b)). Subsequently, the larger fingerprint images acquired from targets in $T_2$ achieved higher match scores against SD4 images than fingerprint images acquired from targets in $T_1$. This finding underscores one of the key advantages of contactless fingerprint readers. In particular, it shows that contactless readers are robust to small mechanical variations in human finger epidermis.

\section{Experiments}

With the fidelity and reproducibility of the universal fingerprint target fabrication process established, multiple experiments are performed on all three major types of fingerprint readers using universal fingerprint targets as operational evaluation targets. First, three fingerprint readers ({\it COR\_A,} {\it CLOR}, and {\it CPR\_A} (Table 5)) are individually assessed using three different universal fingerprint targets mapped with controlled calibration patterns (horizontal gratings, vertical gratings, and circular gratings). Next, the same three fingerprint readers are individually evaluated using impressions acquired from fingerprint targets in $T_2$. Finally, a fingerprint reader interoperability study is performed by comparing images acquired from one of three reader types (contact-optical, contactless-optical, and capacitive) against images acquired from another of the three reader types. 

\begin{table}[b]
 \centering
 \begin{threeparttable}
\caption{Mean ($\mu$) and std. deviation ($\sigma$) of center-to-center ridge spacings (in pixels) on images acquired from 3 universal fingerprint targets. The expected ridge spacing is 9.8 pixels. }
\begin{tabular}{|c||c|c|c|}
 \hline
 \specialcell{Sine Gratings \\Pattern} & \specialcell{Contact \\Optical\\({\it COR\_A})} & \specialcell{Contactless \\ Optical \\({\it CLOR})} & \specialcell{Capacitive\\({\it CPR\_A})}  \\
 \hline
 \hline
 Circular & \specialcell{$\mu = 9.50$ \\$\sigma = 0.56$} & \specialcell{$\mu = 8.99$ \\$\sigma = 0.06$} & \specialcell{$\mu = 9.75$ \\$\sigma= 0.12$}\\
 \hline
Horizontal & \specialcell{$\mu = 9.21$ \\$\sigma = 0.65$} & \specialcell{$\mu = 8.94$ \\$\sigma = 0.16$} & \specialcell{$\mu = 9.45$ \\$\sigma= 0.10$}\\
  \hline
Vertical & \specialcell{$\mu = 8.90$ \\$\sigma = 0.88$} & \specialcell{$\mu = 7.63$ \\$\sigma = 0.51$} & \specialcell{$\mu = 9.17$ \\$\sigma= 0.09$}\\
  \hline
\end{tabular}
\end{threeparttable}
\vspace{-2.0em}
\end{table}

\newcommand{\addpic}{\includegraphics[width=8em]{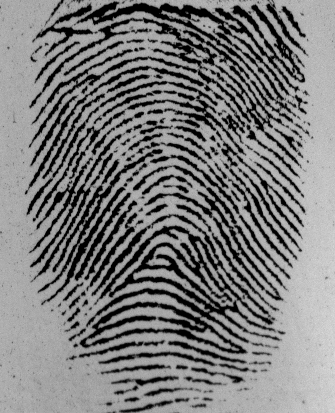}}
\newcommand{\addpictwo}{\includegraphics[width=8em]{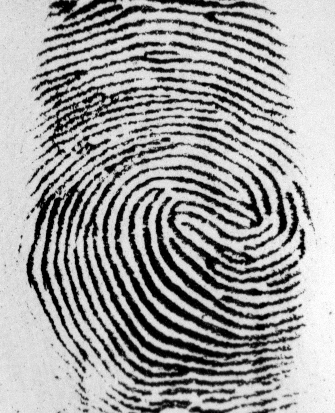}}
\newcommand{\addpicthree}{\includegraphics[width=8em]{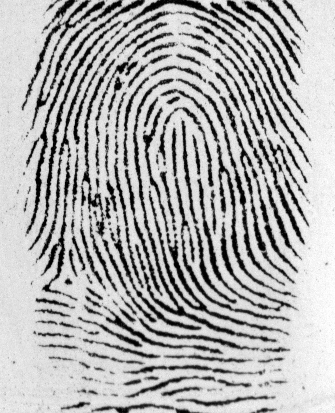}}
\newcommand{\addpicfour}{\includegraphics[width=8em]{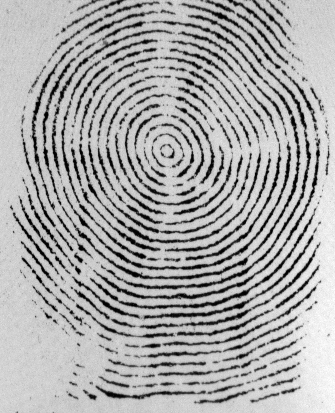}}
\newcommand{\addpicfive}{\includegraphics[width=8em]{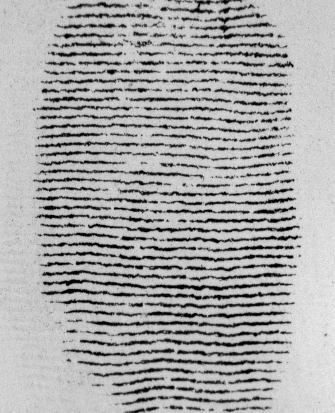}}
\newcommand{\addpicsix}{\includegraphics[width=8em]{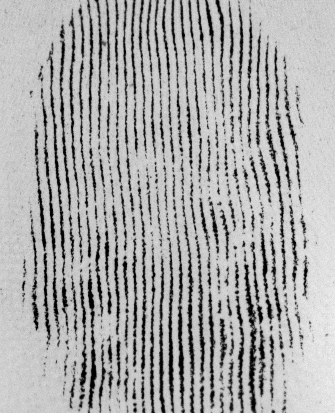}}

\newcommand{\addpicseven}{\includegraphics[width=8em]{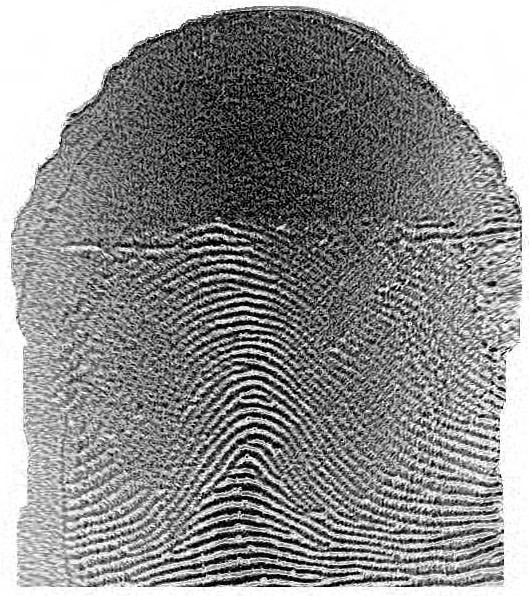}}
\newcommand{\addpiceight}{\includegraphics[width=8em]{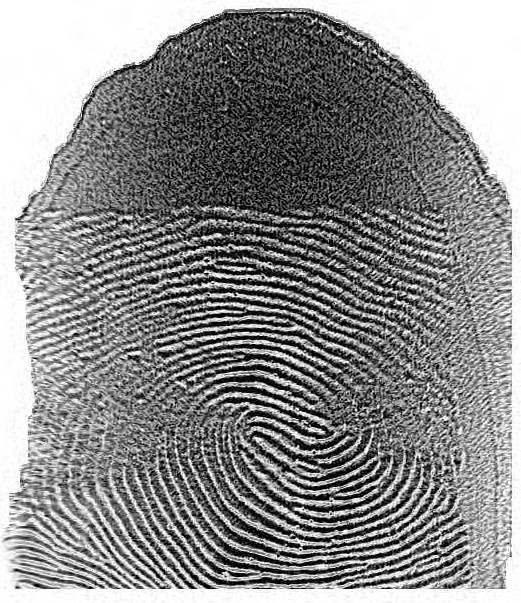}}
\newcommand{\addpicnine}{\includegraphics[width=8em]{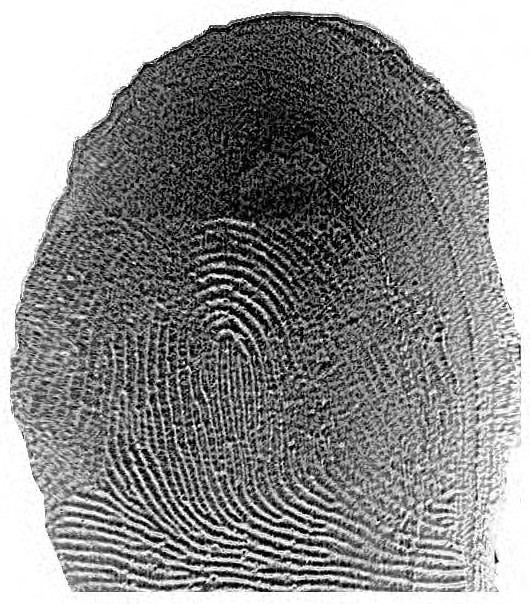}}
\newcommand{\addpicten}{\includegraphics[width=8em]{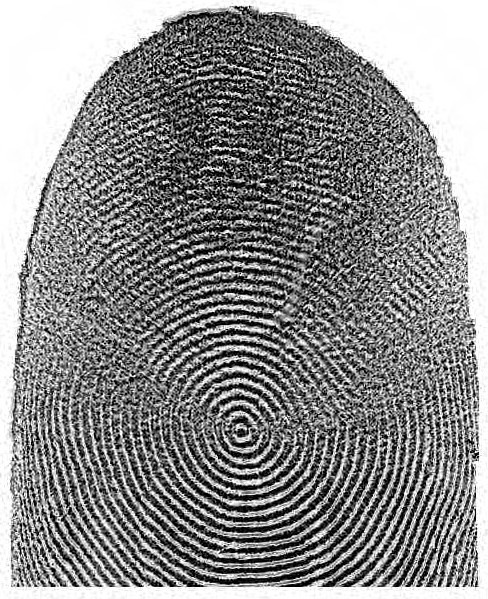}}
\newcommand{\addpiceleven}{\includegraphics[width=8em]{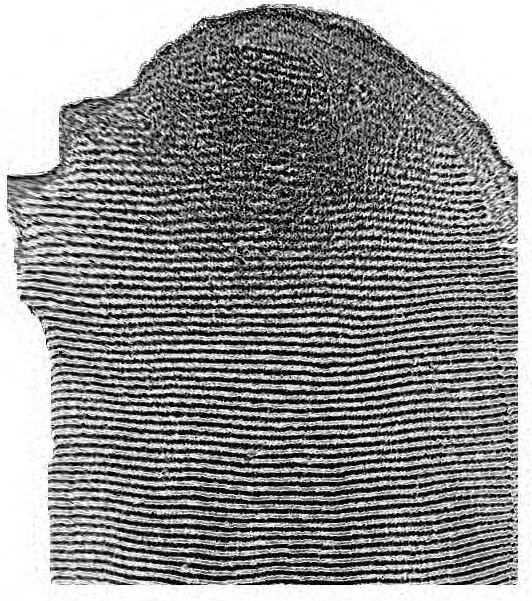}}
\newcommand{\addpictwelve}{\includegraphics[width=8em]{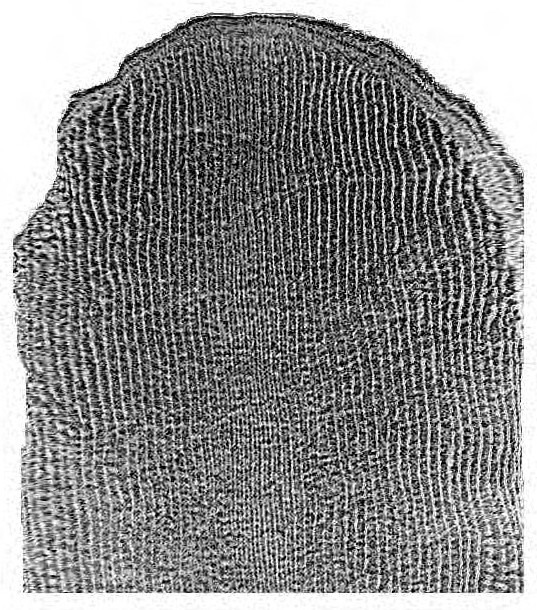}}

\newcommand{\addpicthirt}{\includegraphics[width=8em]{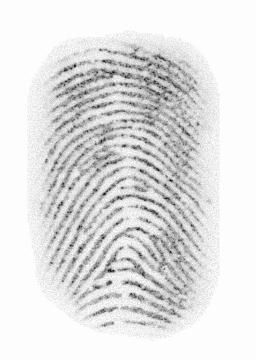}}
\newcommand{\addpicfourt}{\includegraphics[width=8em]{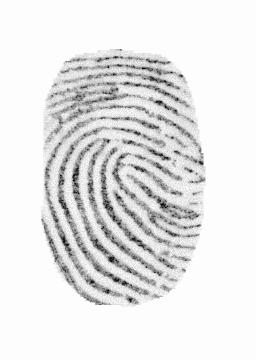}}
\newcommand{\addpicfift}{\includegraphics[width=8em]{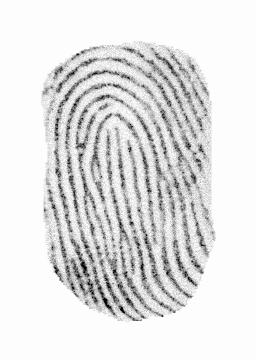}}
\newcommand{\addpicsixt}{\includegraphics[width=8em]{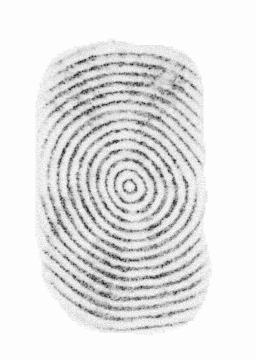}}
\newcommand{\addpicsevent}{\includegraphics[width=8em]{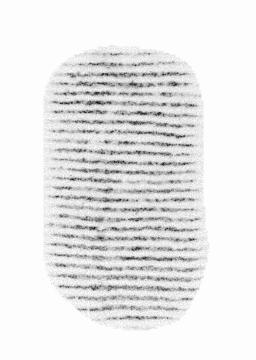}}
\newcommand{\addpiceightt}{\includegraphics[width=8em]{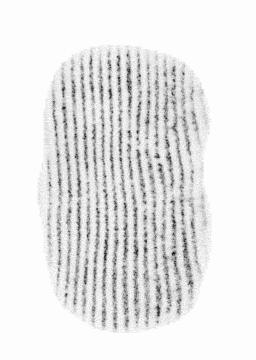}}

\newcolumntype{C}{>{\centering\arraybackslash}m{6.25em}}
\begin{figure*}\sffamily
\begin{tabular}{l*6{C}@{}}
\toprule
\specialcell{Reader Type} & S0005 & S0010 & S0083 & Circular Gratings & Horizontal Gratings & Vertical Gratings \\ 
\midrule
\specialcell{Contact\\Optical} & \addpic & \addpictwo & \addpicthree & \addpicfour & \addpicfive & \addpicsix \\ 

\specialcell{Contactless\\Optical} & \addpicseven & \addpiceight & \addpicnine & \addpicten & \addpiceleven & \addpictwelve \\ 
\vspace{-.5em}
\specialcell{Capacitive} & \addpicthirt & \addpicfourt & \addpicfift & \addpicsixt & \addpicsevent & \addpiceightt \\ 
\bottomrule 
\end{tabular}
\caption{Example fingerprint impressions from 6 universal fingerprint targets (one per column) on 3 types of fingerprint readers. }
\vspace{-1.0em}
\end{figure*}

\subsection{Evaluating Readers with Calibration Patterns}

To evaluate the directional imaging capability of fingerprint readers, we design a similar experiment to that which is proposed in \cite{3Dfingers}. In particular, we collect 10 impressions on 3 different types of fingerprint readers using 3 different universal fingerprint targets mapped with controlled calibration patterns (example impressions shown in Fig. 14). Then, using the method in \cite{ridge_count} the average ridge-to-ridge spacing (in pixels) is computed for the captured impressions. Unlike the targets proposed in \cite{3Dfingers, whole_hand, goldfinger} which could only perform directional assessment of one type of fingerprint reader, our proposed universal fingerprint targets are capable of performing directional assessment on contact-optical, contactless-optical, and capacitive fingerprint readers alike. Therefore, in Table 7, we report the average ridge-to-ridge spacing of the 3 different universal fingerprint targets across all three of the major fingerprint reader types.

All three of the calibration patterns that were mapped to universal fingerprint targets have a 10 pixel peak-to-peak frequency. Given our earlier findings of an approximately 2 \% decrease in point-to-point distances on the universal fingerprint targets during fabrication (due to silicone shrinkage), ridge-to-ridge distances on the 3 calibration mapped universal fingerprint targets are expected to be 9.8 pixels. Given this ground truth value and the results of Table 7, we can evaluate the three types of fingerprint readers used in this experiment. 

The summary of our findings are as follows:

\begin{itemize}
\item Similar to the findings of \cite{3Dfingers}, impressions of targets mapped with circular gratings have larger ridge-to-ridge spacing than impressions of targets mapped with horizontal or vertical gratings. As noted in \cite{3Dfingers}, this is likely due to the radial flattening of the target with circular gratings as it is applied with pressure to the fingerprint reader platen. This radial flattening results in larger ridge-to-ridge spacing than the flattening of the horizontal and vertical calibration targets.
\item Unlike the findings of \cite{3Dfingers, goldfinger}, all of the captured impressions of universal fingerprint targets have smaller ridge-to-ridge spacing than the expected ridge-to-ridge spacing. In \cite{3Dfingers, goldfinger} a larger than expected ridge-to-ridge spacing was explained as a result of ridge-to-ridge distance expansion during the flattening of the target against the reader platen. We hypothesize that universal fingerprint targets have smaller ridge-to-ridge expansion during contact with the reader platen than \cite{3Dfingers, goldfinger} since universal fingerprint targets are less elastic than the targets in \cite{3Dfingers, goldfinger}. Universal fingerprint targets are closer in elasticity to the human skin than \cite{3Dfingers, goldfinger} and so the results shown in Table 7 are more indicative of the ridge-to-ridge spacing the readers used in this study are able to capture from real human fingers.
\item Consistent with the findings of \cite{whole_hand}, the ridge-to-ridge distances are smaller on the contactless fingerprint reader than on the contact fingerprint readers. In particular, the captured ridge-to-ridge spacing of the vertical gratings was lower than expected. We hypothesize that the ridge-to-ridge spacing on the contactless reader is smaller due to the fact that no distortion occurs during image acquisition (as no pressure is applied onto a reader platen). Further analysis needs to be undertaken to understand why the vertical gratings deviated most from the expected ridge spacing. 


\end{itemize}

\subsection{Evaluating Readers with Fingerprint Patterns}

Similar to the experiment in 6.1, we conduct an analysis of the ridge-to-ridge distances captured by three of the major fingerprint reader types. However, in this experiment, rather than mapping controlled calibration patterns to universal fingerprint targets, we use the targets from $T_2$ which are each mapped with real fingerprint images from SD4. In doing so, we evaluate the readers with targets very similar to the real fingers the readers will see in an operational setting.

\begin{table}[!b]
 \centering
 \begin{threeparttable}
\caption{Mean ($\mu$) and std. deviation ($\sigma$) of center-to-center ridge spacings (in pixels) on images acquired from 6 universal fingerprint targets. Expected ridge spacing (in pixels) for each target is reported in parenthesis}
\begin{tabular}{|c||c|c|c|}
 \hline
 \specialcell{SD4 Fingerprint} & \specialcell{Contact \\Optical \\({\it COR\_A)}} & \specialcell{Contactless\\ Optical \\({\it CLOR})} & \specialcell{Capacitive \\({\it CPR\_A})}  \\
 \hline
 \hline
 S0005 (9.25 )& \specialcell{$\mu = 8.77$ \\$\sigma = 1.17$} & \specialcell{$\mu = 8.77$ \\$\sigma = 0.31$} & \specialcell{$\mu = 9.01$ \\$\sigma= 0.18$}\\
 \hline
S0010 (9.98) & \specialcell{$\mu = 9.87$ \\$\sigma = 1.46$} & \specialcell{$\mu = 9.52$ \\$\sigma = 0.29$} & \specialcell{$\mu = 10.42$ \\$\sigma= 0.41$}\\
  \hline
S0031 (10.37) & \specialcell{$\mu = 10.02$ \\$\sigma = 1.40$} & \specialcell{$\mu = 9.04$ \\$\sigma = 0.37$} & \specialcell{$\mu = 10.45$ \\$\sigma= 0.28$}\\
  \hline
 S0044 (9.07) & \specialcell{$\mu = 8.49$ \\$\sigma = 1.24$} & \specialcell{$\mu = 8.25$ \\$\sigma = 0.18$} & \specialcell{$\mu = 9.04$ \\$\sigma= 0.23$}\\
  \hline
S0068 (9.48) & \specialcell{$\mu = 9.60$ \\$\sigma =  1.29$} & \specialcell{$\mu = 9.18$ \\$\sigma = 0.29$} & \specialcell{$\mu = 9.86$ \\$\sigma= 0.19$}\\
  \hline
S0083 (10.23) & \specialcell{$\mu = 9.70$ \\$\sigma = 1.23$} & \specialcell{$\mu = 8.16$ \\$\sigma = 0.15$} & \specialcell{$\mu = 10.23$ \\$\sigma= 0.14$}\\
  \hline
\end{tabular}
\end{threeparttable}
\end{table}

 \begin{table*}[t]
 \centering
 \caption{Genuine and Imposter Score\textsuperscript{1} Statistics and Matching Performance Measures when Comparing Fingerprint Images Acquired from Different Types of Fingerprint Readers. Mean of Genuine Scores ($\mu G$), Mean of Imposter Scores ($\mu I$), True Accept Rate (TAR) and False Accept Rate (FAR) are Reported.}
\begin{threeparttable}
\resizebox{\textwidth}{!}{
\begin{tabular}{ |c||c|c|c|c|c|}
\hline
 & \multicolumn{5}{c|}{\specialcell{Probe Image \\Fingerprint Readers}}\\
 \hline
\specialcell{Enrollment Image \\Fingerprint Readers} & \specialcell{Contact-Optical \\{\it COR\_A}} &\specialcell{Contact-Optical \\{\it COR\_B}}&\specialcell{Contactless-Optical \\{\it CLOR}}& \specialcell{Capacitive \\{\it CPR\_A}} &\specialcell{Capacitive \\{\it CPR\_B}} \\
\hline
\hline
 \specialcell{Contact-Optical \\ {\it COR\_A} } & \specialcell{$\mu G = 440.69$, $\mu I = 0.46$ \\ $TAR = 100\%$, $FAR = 0.0\%$} & \specialcell{
 $\mu G = 399.63$, $\mu I = 0.33$ \\ $TAR = 100\%$, $FAR = 0.0\%$}   &\specialcell{
 $\mu G = 182.20$, $\mu I = 1.16$ \\ $TAR = 100\%$, $FAR = 0.0\%$}  & \specialcell{
 $\mu G = 275.99$, $\mu I = 1.88$ \\ $TAR = 100\%$, $FAR = 0.0\%$}   & \specialcell{
 $\mu G = 202.43$, $\mu I = 4.76$ \\ $TAR = 100\%$, $FAR = 0.0\%$}  \\
 \hline
 \specialcell{Contact-Optical \\{\it COR\_B}} & \specialcell{
 $\mu G = 399.32$, $\mu I = 0.28$ \\ $TAR = 100\%$, $FAR = 0.0\%$} & \specialcell{
 $\mu G = 438.06$, $\mu I = 0.17$ \\ $TAR = 100\%$, $FAR = 0.0\%$}   &\specialcell{
 $\mu G = 171.30$, $\mu I = 0.50$ \\ $TAR = 99.83\%$, $FAR = 0.0\%$}  & \specialcell{
 $\mu G = 278.54$, $\mu I = 1.57$ \\ $TAR = 100\%$, $FAR = 0.0\%$}   & \specialcell{
 $\mu G = 200.03$, $\mu I = 4.25$ \\ $TAR = 100\%$, $FAR = 0.0\%$}  \\
  \hline
 \specialcell{Contactless-Optical\\{\it CLOR}} & \specialcell{
 $\mu G = 183.76$, $\mu I = 1.39$ \\ $TAR = 100\%$, $FAR = 0.0\%$} & \specialcell{
 $\mu G = 174.29$, $\mu I = 0.54$ \\ $TAR = 100\%$, $FAR = 0.0\%$}   &\specialcell{
 $\mu G = 334.06$, $\mu I = 8.99$ \\ $TAR = 100\%$, $FAR = 0.0\%$}  & \specialcell{
 $\mu G = 154.06$, $\mu I = 2.11$ \\ $TAR = 99.83\%$, $FAR = 0.0\%$}   & \specialcell{
 $\mu G = 113.20$, $\mu I = 4.59$ \\ $TAR = 94.83\%$, $FAR = 0.0\%$}  \\
  \hline
 \specialcell{Capacitive \\{\it CPR\_A}} & \specialcell{
 $\mu G = 271.07$, $\mu I = 0.83$ \\ $TAR = 100\%$, $FAR = 0.0\%$} & \specialcell{
 $\mu G = 274.71$, $\mu I = 0.83$ \\ $TAR = 100\%$, $FAR = 0.0\%$}   &\specialcell{
 $\mu G = 146.99$, $\mu I = 1.63$ \\ $TAR = 99.67\%$, $FAR = 0.0\%$}  & \specialcell{
 $\mu G = 352.99$, $\mu I = 7.58$ \\ $TAR = 100\%$, $FAR = 0.0\%$}   & \specialcell{
 $\mu G = 269.37$, $\mu I = 12.08$ \\ $TAR = 100\%$, $FAR = 0.0\%$}  \\
  \hline
 \specialcell{Capacitive \\{\it CPR\_B}} & \specialcell{
 $\mu G = 196.40$, $\mu I = 2.26$ \\ $TAR = 100\%$, $FAR = 0.0\%$} & \specialcell{
 $\mu G = 195.38$, $\mu I = 2.24$ \\ $TAR = 100\%$, $FAR = 0.0\%$}   &\specialcell{
 $\mu G = 105.37$, $\mu I = 3.237$ \\ $TAR = 91.83\%$, $FAR = 0.0\%$}  & \specialcell{
 $\mu G = 268.16$, $\mu I = 10.08$ \\ $TAR = 100\%$, $FAR = 0.0\%$}   & \specialcell{
 $\mu G = 277.48$, $\mu I = 14.35$ \\ $TAR = 100\%$, $FAR = 0.0\%$}  \\
  \hline
 
\end{tabular}}
\begin{tablenotes}
\item[1] Innovatrics matcher was used to generate similarity scores. The threshold of the matcher at FAR = 0.01 \% was computed to be 49 on the FVC \\2002 and 2004 databases \cite{fvc2002, fvc2004}.
\end{tablenotes}
\end{threeparttable}
\vspace{-1.0em}
\end{table*}

Again, 10 impressions are captured on all 3 fingerprint readers, this time with each of the 6 universal fingerprint targets in $T_2$ (example impressions shown in Fig. 14). Then, using the method in \cite{ridge_count}, the average ridge spacing of the captured impressions is computed (Table 8). Additionally, the average ridge spacing is computed (using the method in \cite{ridge_count}) on the original fingerprint images from SD4 and established as the ground truth ridge spacing values. By comparing these ground truth values with the results of Table 8, we perform an assessment of the three fingerprint readers. 

In summary, the findings of this experiment are as follows:

\begin{itemize}

\item Consistent with the findings of our experiment in 6.1 with calibration pattern mapped universal fingerprint targets, the images captured by the contactless-optical fingerprint reader have smaller ridge-to-ridge distances than the impressions captured by contact based readers. This is likely due to the absence of fingerprint distortions in contactless fingerprint readers. Additionally, errors in the contactless reader may be introduced when the three-dimensional finger surface captured by the reader is projected into two dimensions (due to the ridge height of universal fingerprint targets being greater than the ridge height of human fingers). 
\item In almost all of the target impressions, capacitive fingerprint readers captured the ridge-to-ridge distances more closely to ground truth than contact-optical readers did. Further studies and analysis need to be performed to determine if this finding is consistent, and also, the explanation behind this.
\end{itemize}

\subsection{Reader Interoperability Evaluations}
Whereas our previous two experiments in 6.1 and 6.2 evaluated the three major types of fingerprint readers individually, in this final experiment, we perform fingerprint reader {\it interoperability} evaluations using the universal fingerprint targets.

To set up this experiment, 10 impressions from each target in $T_2$ are captured on 5 different fingerprint readers (Table 5). Then, for all pairs of fingerprint readers in our set of 5 readers, images from one reader are used as enrollment images and images from the other reader are used as probe images to generate genuine and imposter scores using the Innovatrics matcher \cite{in}. In Table 9, we report the means of the genuine and imposter scores. Additionally we report the True Accept Rate (TAR) and the False Accept Rate (FAR) of the scores using a threshold of 49 (this threshold was precomputed on the FVC 2002 and 2004 databases \cite{fvc2002, fvc2004}, because we do not have a sufficient number of images from the targets to set the threshold). 

Although the performance results of Table 9 seem to indicate that all of the readers used are highly interoperable, these results are likely too optimistic as only 6 different targets were used. For this reason, we also report the genuine and imposter score means to show how the scores deteriorate when different readers are used for enrollment and verification. Similar to the findings of past fingerprint reader interoperability studies \cite{Ross2004, ross_inter, cost_inter}, we note that genuine scores decrease and imposter scores increase when different fingerprint readers are used to acquire enrollment images and probe images, especially when the two readers use different sensing technology to acquire images. While past studies reported these findings using real fingers for data collection, we report the same findings, for the first time ever, using realistic, 3D, wearable, fingerprint targets. By demonstrating the same results as past studies with the universal fingerprint targets, we validate the utility in using universal fingerprint targets for advancing fingerprint reader interoperability studies. In particular, the universal fingerprint targets could be mounted to a robot and imaged on different readers at known pressure and orientation. This standardized data could then be used to learn calibration mappings between different fingerprint readers which could be used to drastically improve fingerprint reader interoperability.

\section{Conclusions and Future Work}
We have designed a molding and casting system capable of fabricating wearable, 3D fingerprint targets from a plethora of casting materials. By selecting a casting material with similar mechanical, optical, and electrical properties to the human skin, we cast {\it universal fingerprint targets}, which can be imaged on the three major fingerprint reader types in use (contact-optical, contactless-optical, and capacitive). Previous studies were unable to produce a single 3D fingerprint target which could be imaged on multiple types of fingerprint readers. We demonstrate that the process for fabricating universal fingerprint targets is of high fidelity, and that it is reproducible. Finally, we use the universal fingerprint targets as evaluation targets on multiple types of PIV/Appendix F certified fingerprint readers. Our results verify the utility in using the universal fingerprint targets for both individual fingerprint reader assessments and also fingerprint reader interoperability studies. We believe that the universal 3D fingerprint targets introduced here will advance state of the art in fingerprint reader evaluation and interoperability studies.

In the future, the universal fingerprint targets will be mounted to a robotic hand and imaged on various fingerprint readers at known pressure and orientation. With this data, objective evaluations can be performed on fingerprint readers. Additionally, the data collected could be utilized to learn fingerprint distortion models, fingerprint reader interoperability calibration models, and latent fingerprint distortion models. Finally, the universal fingerprint targets will be used to assess the spoofing vulnerability of various fingerprint recognition systems (such as smartphones).


%

\ifCLASSOPTIONcompsoc
  \section*{Acknowledgments}
\else
  \section*{Acknowledgment}
\fi

This research was supported by grant no. 60NANB11D155
from the NIST Measurement Science program. The authors would like to thank Brian Wright, Michigan State University, for his help in 3D printing of molds. We would also like to thank Edward Drown, Michigan State University, for his help mixing castings materials for the universal fingerprint targets.

\ifCLASSOPTIONcaptionsoff
  \newpage
\fi



%
\bibliography{cites}
\bibliographystyle{ieeetr}

%
\vspace{-15.0 mm}
\begin{IEEEbiography}[{\includegraphics[width=1in,height=1.25in,clip,keepaspectratio]{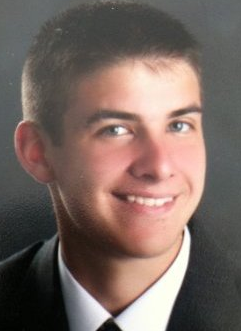}}]{Joshua J. Engelsma}
received his B.S. degree
in computer science from Grand Valley State University, 
Allendale, Michigan, in 2016. He is currently
working towards a PhD degree in the
Department of Computer Science and Engineering
at Michigan State University, East Lansing,
Michigan. His research interests include pattern
recognition, computer vision, and image processing
with applications in biometrics.
\end{IEEEbiography}

\vspace{-15.0 mm}
\begin{IEEEbiography}[{\includegraphics[width=1in,height=1.25in,clip,keepaspectratio]{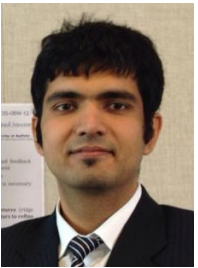}}]{Sunpreet S. Arora}
received the B.Tech. (Hons.)
degree in Computer Science from the Indraprastha
Institute of Information Technology, Delhi (IIIT-D)
in 2012, and the Ph.D. degree in Computer Science
and Engineering from Michigan State University, in 2016. He is currently a Senior Biometrics
Researcher at Visa Inc., Foster City, CA. His
research interests include biometrics, pattern recognition
and image processing. He received the best
paper award at the 15th IEEE BIOSIG,
2016, and the best poster award at the IEEE BTAS, 2012.
\end{IEEEbiography}

\vspace{-15.0 mm}
\begin{IEEEbiography}[{\includegraphics[width=1in,height=1.25in,clip,keepaspectratio]{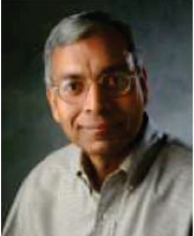}}]{Anil K. Jain}
is a University distinguished professor
in the Department of Computer Science and
Engineering at Michigan State University.
His research interests include pattern recognition and
biometric authentication. He served as the editor-in-chief
of the IEEE Transactions on Pattern Analysis
and Machine Intelligence. He is a member of the United States National Academy
of Engineering and a Foreign Fellow of the Indian National Academy of
Engineering.
\end{IEEEbiography}

\vspace{-15.0 mm}
\begin{IEEEbiography}[{\includegraphics[width=1in,height=1.25in,clip,keepaspectratio]{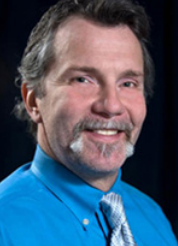}}]{Nicholas G. Paulter Jr.}
is the Group Leader for the
Security Technologies Group at NIST in Gaithersburg,
MD. He develops and oversees metrology programs
related to concealed weapon and contraband
imaging and detection, biometrics for identification,
and body armor characterization. He has authored
or co-authored over 100 peer-reviewed technical
articles and provided numerous presentations at a
variety of technical conferences. He is a 2008-2009
Commerce Science and Technology Fellow and a
2010 IEEE Fellow
\end{IEEEbiography}




\end{document}